\documentclass[acmtog,nonacm]{acmart}
\renewcommand\footnotetextcopyrightpermission[1]{}
\makeatletter
\let\@authorsaddresses\@empty
\makeatother\usepackage{booktabs} 
\usepackage{tablefootnote}

\newcommand{\src}{\mathcal{S}}
\newcommand{\tar}{\mathcal{T}}

\usepackage{multirow}
\usepackage{colortbl}
\usepackage[ruled]{algorithm2e} 

\SetAlFnt{\small}
\SetAlCapFnt{\small}
\SetAlCapNameFnt{\small}
\SetAlCapHSkip{0pt}

\author{Mingze Sun}
\authornote{Equal contribution. }
\affiliation{%
  \institution{Tsinghua Shenzhen International Graduate School}
  \country{China}
}

\author{Chen Guo}
\authornotemark[1]
\affiliation{%
  \institution{Tsinghua Shenzhen International Graduate School, Pengcheng Lab}
  \country{China}
}

\author{Puhua Jiang}
\authornotemark[1]
\affiliation{%
  \institution{Tsinghua Shenzhen International Graduate School, Pengcheng Lab}
    \country{China}
}

\author{Shiwei Mao}
\affiliation{%
  \institution{Tsinghua Shenzhen International Graduate School}
    \country{China}
}

\author{Yurun Chen}
\affiliation{%
  \institution{Tsinghua Shenzhen International Graduate School}
    \country{China}
}

\author{Ruqi Huang}\authornote{Corresponding to Ruqi Huang (ruqihuang@sz.tsinghua.edu.cn).}
\affiliation{%
  \institution{Tsinghua Shenzhen International Graduate School}
    \country{China}
}

\begin{document}
\title{\textbf{SRIF}: \textbf{S}emantic Shape \textbf{R}egistration Empowered by Diffusion-based \textbf{I}mage Morphing and \textbf{F}low Estimation}

\begin{teaserfigure}
  \begin{center}
\includegraphics[width=15cm]{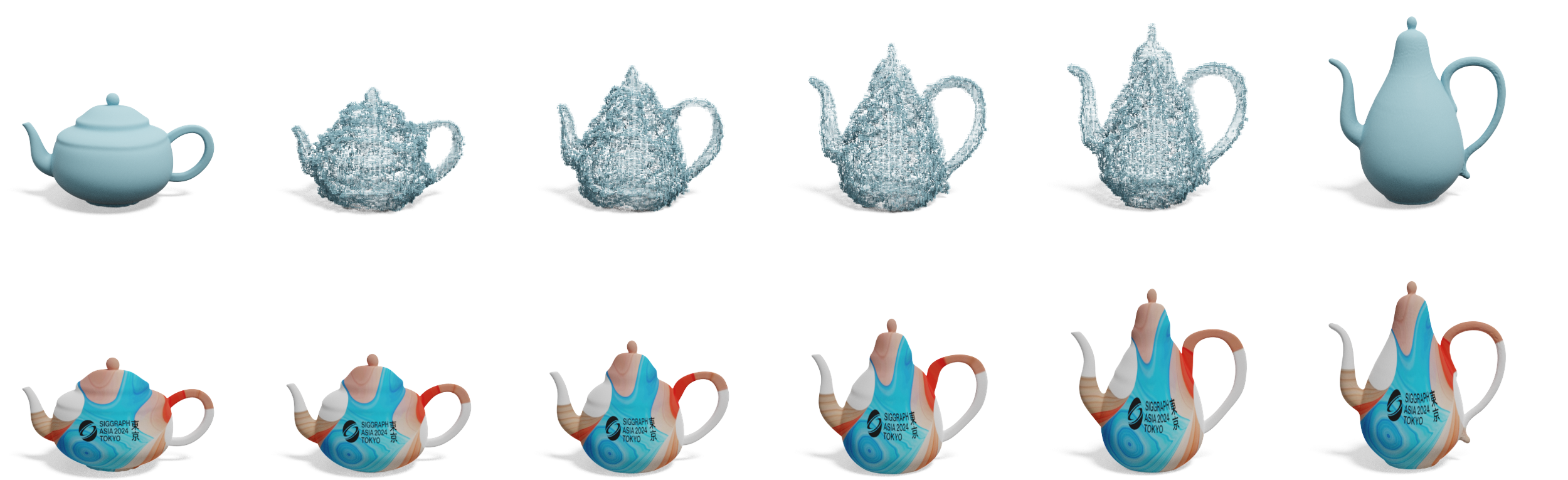}
  \end{center}
\caption{Given two teapots (blue meshes),  \textbf{SRIF} first generates plausible intermediate point clouds (middle of top row) bridging them based on multi-view image  morphing~\cite{diffmorpher} and dynamic 3D Gaussian splatting reconstruction~\cite{sc-gs}, and then estimates an invertible normalizing flow that \emph{continuously} deforms the source to the target with the above auxiliary point clouds, resulting in both a semantically meaningful morphing process and high-quality dense correspondences indicated by the accurate texture transfer (bottom row).   }
\label{fig:teaser}
\end{teaserfigure}

\begin{abstract}

In this paper, we propose \textbf{SRIF}, a novel \textbf{S}emantic shape \textbf{R}egistration framework based on diffusion-based \textbf{I}mage morphing and \textbf{F}low estimation. 
More concretely, given a pair of extrinsically aligned shapes, we first render them from multi-views, and then utilize an image interpolation framework based on diffusion models to generate sequences of intermediate images between them. 
The images are later fed into a dynamic 3D Gaussian splatting framework, with which we reconstruct and post-process for intermediate \emph{point clouds} respecting the image morphing processing. 
In the end, tailored for the above, we propose a novel registration module to estimate continuous normalizing flow, which deforms source shape consistently towards the target, with intermediate point clouds as weak guidance. 
Our key insight is to leverage large vision models (LVMs) to \emph{associate} shapes and therefore obtain much richer semantic information on the relationship between shapes than the ad-hoc feature extraction and alignment. 
As consequence, \textbf{SRIF} achieves high-quality dense correspondences
on challenging shape pairs, but also delivers smooth, semantically
meaningful interpolation in between.
Empirical evidences justify the effectiveness and superiority of our method as well as specific design choices. 
The code is released at \url{https://github.com/rqhuang88/SRIF}. 

\end{abstract}



\maketitle

\section{Introduction}\label{sec:intro}

Estimating dense correspondences between 3D shapes serves as a cornerstone in many applications of computer graphics, including 3D reconstruction~\cite{yu2018doublefusion}, animation~\cite{sumner2004deformation} and statistical shape analysis~\cite{scape}, to name a few. 
Regarding shapes undergoing rigid or isometric deformations, the prior shape registration/matching techniques~\cite{icp, nicp, gmds, ovsjanikov2012functional} have laid down solid foundations on both theoretical and practical fronts. 
In this paper, we consider the problem of estimating semantically meaningful dense correspondences between shapes undergoing more general and complicated deformations.  

In the absence of a compact deformation prior, the purely geometric methods typically take a coarse-to-fine approach. 
Namely, one leverages geometric features to locate a small set of landmarks on both shapes, estimates sparse landmark correspondences, and finally propagates dense correspondences via minimizing distortions such as conformal~\cite{kim2011blended}, elastic energy~\cite{enigma}. 
It is worth noting, though, that the sparse correspondences derived from geometry are not necessarily relevant to semantics. 
As shown in qualitative results in Sec.~\ref{sec:exp}, the resulting maps can suffer from such discrepancy, especially in the presence of significant heterogeneity.

\begin{figure}[t!]
  \begin{center}
\includegraphics[width=8.5cm]{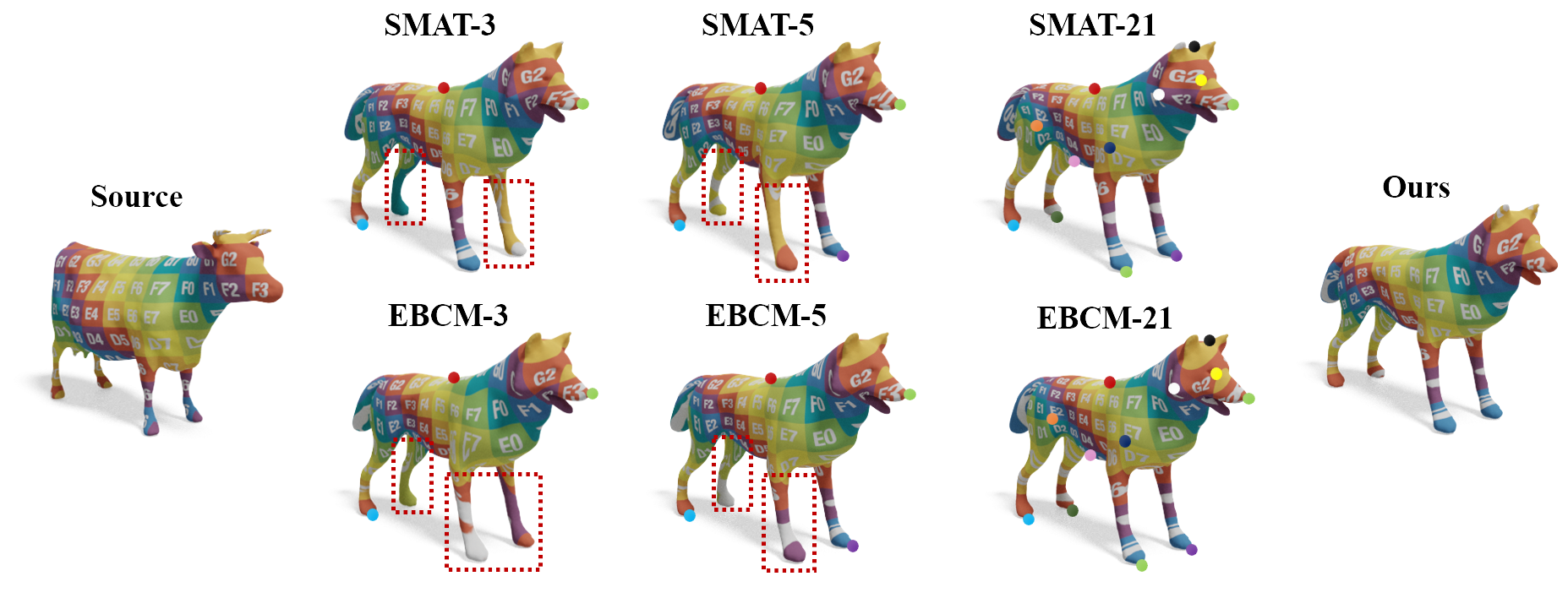}
  \end{center}
  \vspace{-5mm}
\caption{Maps obtained by state-of-the-art compatible remeshing techniques~\cite{EBCR,EG} based on varying number of landmark correspondences. The absence of landmarks at local region can lead to erroneous matching (see the red boxes). On the other hand, our method delivers high-quality maps with \emph{no} landmark annotation.  }
\label{fig:landmark}
\end{figure}

To this end, another line of works concentrate on producing high-quality semantic correspondences at the cost of dependency on user-defined landmarks~\cite{ssm, inters, rhm, EG, EBCR}. 
Treating landmarks as anchor points, they cast the problem as correspondence interpolation, which can be conducted by optimizing certain geometric distortions.
However, the dependency on manual annotations limits the practical utility of these works. 
Apart from hindering automation, insufficient annotation strength can lead to sub-optimal results as shown in Fig.~\ref{fig:landmark}. 

More recently, the emergence of learning-based techniques has enabled data-driven semantic information extraction.  
In essence, one can learn canonical structural information from a collection of 3D shapes, which can help to match challenging non-rigid shapes~\cite{arapreg, eisenberger2021neuromorph}, or heterogeneous man-made objects~\cite{dif, dit}. 
Unfortunately, limited by the amount of available 3D data, the above approaches are typically \emph{category-specific}, weakening their practical utility. 

On the other hand, boosted by large-scale natural image datasets such as LAION~\cite{schuhmann2022laion5b}, large vision (or vision-language) models (LVMs) ~\cite{dinov2,latentdiffusion, clip, gpt4} has attracted considerable attention from the community of 3D shape analysis ~\cite{satr,zsc, snm, back3d}. 
The shared protocol is to project 3D shapes into multi-view 2D images. 
The latter can then be fed into LVMs for semantic encoding, which is finally aggregated back to 3D shapes for the respective task. 
While the above approach has been proven simple yet effective, we observe that 1) the distilled semantic information is in general coarse (\emph{e.g., }segmented regions or sparse landmarks); 2) there exists a clear gap between multi-view images rendered from textureless 3D shapes and natural images on which LVMs are trained, the feature extracted from LVMs can be noisy; 3) many approaches leverage the semantics in a simple feed-forward manner and therefore require complicated filtering schemes to ensure reliability~\cite{snm}.

\begin{figure*}[t!]
  \begin{center}
\includegraphics[width=15cm]{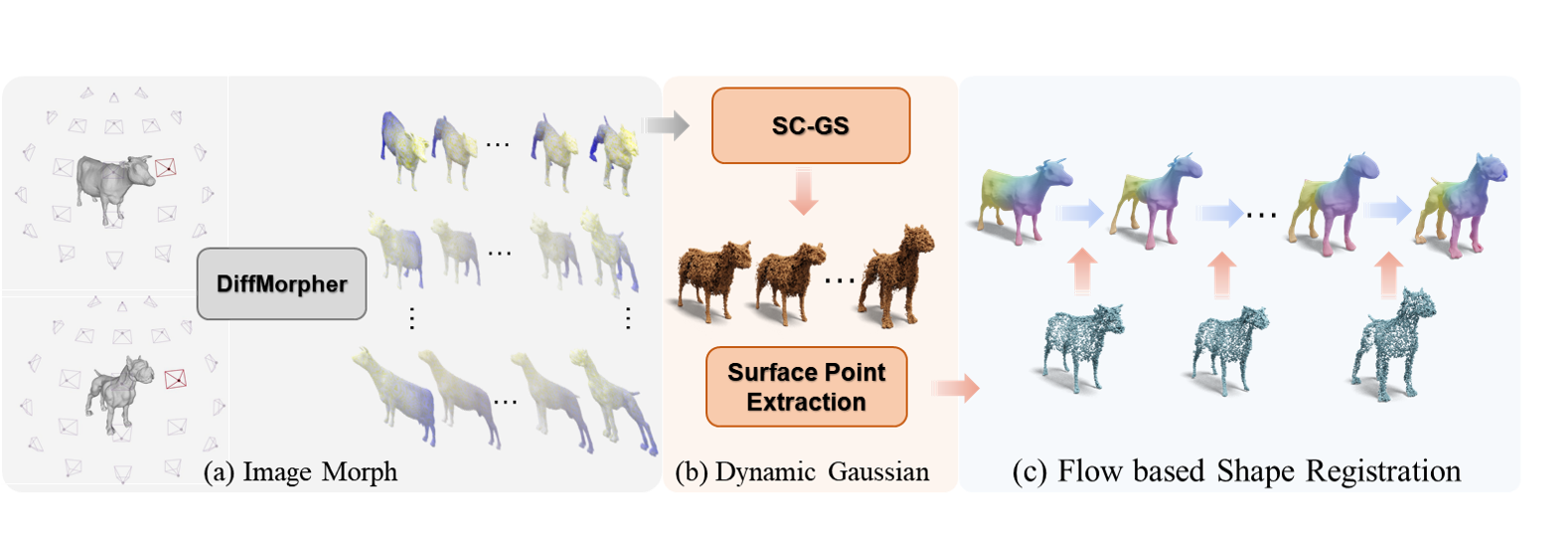}
  \end{center}
  \vspace{-5mm}
\caption{Our pipeline mainly contains three blocks. (a) Given two extrinsically aligned shapes, we first render them from multi-views, and then use DiffMorpher~\cite{diffmorpher} to generate image interpolation with respect to each views. (b) We then use SC-GS 3D Gaussian Splatting~\cite{yang2023deformable} to reconstruct, from which we obtain a set of dense and noisy point clouds (in gold). They are fed into our surface point extraction module to obtain clean intermediate point clouds (in blue). (c) Finally, we estimate a continuous normalizing flow, represented by an MLP, that deforms the source to the target under the guidance of the extracted intermediate point clouds (in blue).  }
\label{fig:pipeline}
\end{figure*}

We believe the above issues originate from the fact that LVMs are used to extract features from shapes in a \emph{static, independent} manner, which can be problematic when the shapes are distinct from each other. 
In light of this, we propose \textbf{SRIF}, a framework for \textbf{S}emantic shape \textbf{R}egistration, which is built with diffusion-based \textbf{I}mage morphing and \textbf{F}low estimation. 
The key insight of \textbf{SRIF} is to take 
a \emph{dynamic} viewpoint of leveraging LVMs to associate 3D shapes. 
More specifically, given a pair of extrinsically aligned shapes, we first render
them from multi-views, then we utilize a diffusion-based image interpolation framework~\cite{diffmorpher} to generate sequences of intermediate images between corresponding views. 
Then we reconstruct the intermediate point clouds from the interpolated images via a dynamic 3D Gaussian splatting reconstruction framework~\cite{sc-gs} and our surface point extractor tailored for the reconstructed Gaussians. 
Note that, with the above procedures, we have not achieved any \emph{explicit} connection between the input shapes. 
Nevertheless, the intermediate point clouds carry rich information on how the shapes are associated from the viewpoint of LVMs. 
Last but not least, inspired by the dynamic nature of the extracted semantics, we formulate the shape registration problem as estimating a \emph{flow} that deforms source shape towards the target, with the intermediate point clouds as guidance. 
In particular, we adopt the framework of PointFlow~\cite{yang2019pointflow} into our pipeline and learn a Multi Layer Perception (MLP) to represent as well as optimize for the flow. 

We extensively evaluate \textbf{SRIF} on a wide range of shape pairs from SHREC'07 dataset~\cite{shrec07} and EBCM~\cite{EBCR}. Empirical evidence demonstrates that \textbf{SRIF} outperforms the competing baselines in all test sets. 
As shown in Fig.~\ref{fig:teaser} and Sec.~\ref{sec:exp}, \textbf{SRIF} not only delivers high-quality dense correspondences between shapes but also generates a continuous, semantically meaningful morphing process, which can potentially contribute to 3D data accumulation.

\vspace{-4mm}
\section{Related Works}

\subsection{Dense Shape Correspondence Estimation}
Since estimating 3D shape correspondence is an extensively studied area, we refer readers to~\cite{tam2013registration} for a comprehensive survey and focus on the most relevant methods, which are autonomous methods for general shape matching or registration.  

\noindent{\textbf{Axiomatic methods }}typically follow a coarse-to-fine manner, which depends on sparse landmark correspondences to achieve dense ones. 
The typical approaches~\cite{kim2011blended, enigma} first extract and match landmarks using geometric features, and then independently or jointly optimize for dense correspondences based on certain distortion quantity, such as conformality, smoothness. 
On the other hand, there exists a line of works leveraging fuzzy correspondences~\cite{ovsjanikov2012functional} to alleviate under-constrained space of dense maps. For instance, MapTree~\cite{maptree} exploits the space of functional maps, SmoothShells~\cite{smoothshells} jointly estimate registration transform in both spatial and functional space. 
As mentioned before, correspondences derived from geometry do not always align with semantics, especially in the presence of heterogeneity. 
Our method enjoys the semantic information from LVMs for accurate correspondence estimation.

\noindent{\textbf{Learning-based methods }} take advantage of prior knowledge extracted by networks. 
According to the sources of prior knowledge, we further classify them as follows:

1) \textit{Large Vision Models (LVMs)}: 
SATR~\cite{satr} renders 3D shapes into multi-view images, gathers coarse semantic labels as part segmentation from each view, and finally back-projects to 3D shapes. 
Similarly to the axiomatic methods, such cues can be further post-processed with off-the-shelf methods like functional maps~\cite{ovsjanikov2012functional} to achieve dense correspondences~\cite{zsc}. 
Since the linguistic signal is not suitable for point-level description, the above approaches suffer from coarse external semantics. 
It is then highly non-trivial to perform fine-grained shape analysis tasks, including registration. 
NSSM~\cite{snm} leverages features from Dinov2~\cite{dinov2} to identify sparse landmark correspondences. 
Being a two-stage method, NSSM can be fragile regarding the mismatched landmarks. On the other hand, as shown in Fig.~\ref{fig:interpolation3row}, \textbf{SRIF} is robust with poor semantic information from LVMs. 

2) \textit{Category-specific Training} is prevailing in learning-based 3D shape analysis. 
We identify the most relevant approach along this line as NeuroMorph~\cite{eisenberger2021neuromorph}, which jointly learns to interpolate and perform registration between shapes. 
However, NeuroMorph requires pre-training on a small set of 3D shapes, which belongs to the same category as the test ones. 
Moreover, the interpolation produced by NeuroMorph is dominated by geometric cues, rather than semantics. 
NeuroMorph is related to our method in the sense that it delivers interpolation as a by-product as well. 
However, our method generates plausible interpolations \emph{before} estimating correspondences, while NeuroMorph jointly optimizes for both. 
As suggested by the experimental results, our scheme makes full use of LVMs and surpasses NeuroMorph by a large margin.

\subsection{Diffusion Model}
Diffusion models~\cite{ddpm, latentdiffusion} have gained significant popularity recently, thanks to their impressive capacity of learning data distribution from large-scale image datasets. 
Some recent works~\cite{zhao2023uni, latentdiffusion, blattmann2023align} have attempted to control the generated results and improve the quality of generation. 
On the other hand, the application of diffusion models in image interpolation receives relatively less attention. During the interpolation process using 2D diffusion models, there is a greater focus on the style of the images rather than on deformation. 
Want et al.~\cite{wang2023interpolating} attempt to interpolate in the latent space of diffusion, but the resulting method suffers from poor generalization capability. 
On the other hand, DiffMorpher~\cite{diffmorpher} has achieved smooth interpolation based on StableDiffusion. 
We therefore exploit it in our pipeline as a tool for image morphing, which returns plausible intermediate images encoding the deformations between the source and the target shape.

\subsection{3D Gaussian Splatting}
As a scene representation, 3D Gaussian Splatting (3D-GS)~\cite{3dgs} represents a 3D scene as a mixture of Gaussian distribution. 
3D-GS has dramatically advanced novel view synthesis (NVS) in terms of accuracy and efficiency. 
Numerous follow-ups have been proposed on 3D-GS, ranging across dynamic NVS~\cite{yang2023deformable, sc-gs}, SLAM~\cite{slam}, and geometry recovery~\cite{meshextr}, to name a few.

In particular, our method directly leverages SC-GS~\cite{sc-gs} to reconstruct intermediate geometry from the morphed images. 
Our point cloud extraction (Sec.~\ref{sec:dgaussian}) is similar to mesh extraction from 3D-GS~\cite{meshextr}, which remains a challenging open problem. 

\section{Methodology}\label{sec:method}

\textbf{SRIF} takes as input a pair of meshes $(\src,\tar)$, which are extrinsically aligned, \emph{i.e.,} roughly in the same up-down and front-back orientation. 
The desired output is a registered source shape $\hat{\src}$ that admits the same triangulation as $\src$ and approximates $\tar$ in geometry. 
We demonstrate our whole pipeline in Fig.~\ref{fig:pipeline}, which consists of Image Rendering and Morphing (Sec.~\ref{sec:render}), Intermediate Point Clouds Reconstruction (Sec.~\ref{sec:dgaussian}), and Flow Estimation (Sec.~\ref{sec:regi}).

\subsection{Image Rendering and Morphing}\label{sec:render}
The key sub-goal of our method is to infer an intermediate morphing process between input shapes. 
Our first step is to employ a diffusion-based image morphing technique, DiffMorpher~\cite{diffmorpher} on multi-view image sets, rendered to both $\src$ and $\tar$.

Specifically, we pre-process the input shapes such that they are centered around the origin and scaled to be inside a unit sphere. 
For $\src$ (resp. $\tar$), we render $K$ views, where the viewpoints are sampled uniformly around the shape in the sphere space with a radius of 3.5 length units.  
We use the renderer from Open3D library~\cite{open3d}. 
We observe that properly endowing texture to shapes plays a critical role in the follow-up image morphing step. 
That is because diffusion models~\cite{latentdiffusion} are generally trained on realistic images, which are distinctive from straightforward renderings of textureless shapes. 
To this end, we explore various coloring schemes for rendering shapes and settle down at the following, which integrates spatial coordinate information (see more details in the Supp. Mat.). 
Given a shape, we denote by $z_{\mbox{max}}, z_{\mbox{min}}$ respectively the maximal and minimal value of the $z-$coordinates of its points, and formulate function $[C_R, C_G, C_B] = [ \lfloor \frac{d}{256}\rfloor, \lfloor \frac{d}{256}\rfloor, d - 256\times \lfloor \frac{d}{256}\rfloor]$ to assign color to each point $(x, y, z)$, where $d = \lfloor \frac{z-z_{\mbox{min}}}{z_{\mbox{max}} - z_{\mbox{min}}}\times 65535 \rfloor$ and $C_R, C_G, C_B$ are respective the intensity of the $RGB$ channels. 
The rendered images can be represented as a set of image pairs: $\mathcal{C}_r = \{(I^{\src}_{i},I^{\tar}_{i})| i = 1,2,...,K\}$.
Subsequently, $\mathcal{C}_r$ is processed through an image morphing algorithm DiffMorpher~\cite{diffmorpher}, which leverages diffusion models to generate $K$ sequences of intermediate images. 
This technique allows for a more nuanced and continuous transformation between corresponding views within each image pair in $\mathcal{C}_r$. 
For each pair, DiffMorpher generates $T$ intermediate images, transitioning from $I^{\src}_{i}$ to $I^{\tar}_{i}$. Consequently, this process yields $\mathcal{C} = \{I^{j}_{i} \mid i = 1,2,\ldots,K; j =1,2,\ldots,T\}$, a comprehensive image set that captures a wide array of views and detailed morphing stages.

\subsection{Intermediate Point Clouds Reconstruction and Post-processing}\label{sec:dgaussian}

Note that $\mathcal{C}$ contains images sampled from different views as well as different morphing stages. 
One straightforward way is to reconstruct the intermediate shapes using multi-view reconstruction with images at the same morphing stage. 
However, since the image morphing is performed independently regarding views, one can hardly guarantee multi-view consistency. 

Therefore, we instead formulate the reconstruction as a dynamic one. 
Moreover, for the sake of efficiency and accuracy, we choose the recent art, SC-GS framework~\cite{sc-gs}, for reconstruction. 
Specifically, we use the vertices of the source mesh to initialize the spatial position of each Gaussian (\emph{i.e., }mean). 
From these points, SC-GS creates a set of 3D Gaussians $G( x , r , s , \sigma)$
defined by a center position $x$, opacity $\sigma$, and 3D covariance matrix $\Sigma$ obtained from quaternion $r$ and scaling $s$.  
For a morphing stage $t\in\{1, 2, \cdots, T\}$, SC-GS takes the positions $x$ as input and predict the $(\delta x,\delta r,\delta s)$. 
Subsequently, the deformed 3D Gaussians $G( x +\delta x , r +\delta r , s +\delta s , \sigma)$
is optimized by the differential Gaussian rasterization pipeline. 
Once optimization is done, given a time step $t$, we extract the set of positions as the raw intermediate point clouds $V_{t}^G$. 

\noindent\textbf{Post-processing on $V_{t}^G$:}
First, we deal with outlier points within $V_{t}^G$, which comes from the floating Gaussians in the reconstruction. 
We compute the Euclidean distance between each point and its nearest neighbor in $V_t^G$, and filter out the ones with distances larger than a fixed threshold. 
On the other hand, the adaptive density control module of SC-GS generates additional 3D Gaussians inside the surface of the intermediate shapes. 
The inner points can be misleading for the registration procedure in Sec.~\ref{sec:regi}. 
To accurately delineate these points, we propose a surface point extraction module. 
To be precise, given $V_{t}^G$, we project depth maps from each facet of a standardized hexahedron. 
These depth maps are subsequently re-projected as partial viewpoint clouds. 
After aggregating all these point clouds, we finally obtain the surface point cloud  $V_{t}$ for the registration process. 
We refer readers to the Supp. Mat. for more detailed descriptions. 
After the above procedure, there still exists a large amount of points, which can be redundant in the follow-up registration. 
We thus perform Furthest Point Sampling (FPS) on each $V_{t}$ such that its number of points is same as that of $\src$.

\subsection{3D Shape Registration via Flow Estimation}\label{sec:regi}
Going through the above two main components, we obtain a sequence of intermediate point clouds denoted as $\mathcal{V} = \{V_{t} \mid t = 1,...,T \}$, each corresponds to a one-time step in image morphing. 
Without loss of generality, we denote by $V_0$ the vertices of $\src$, and $V_{T+1}$ that of $\tar$. 
\begin{figure*}[htbp!]
  \begin{center}
\includegraphics[width=17.5cm]{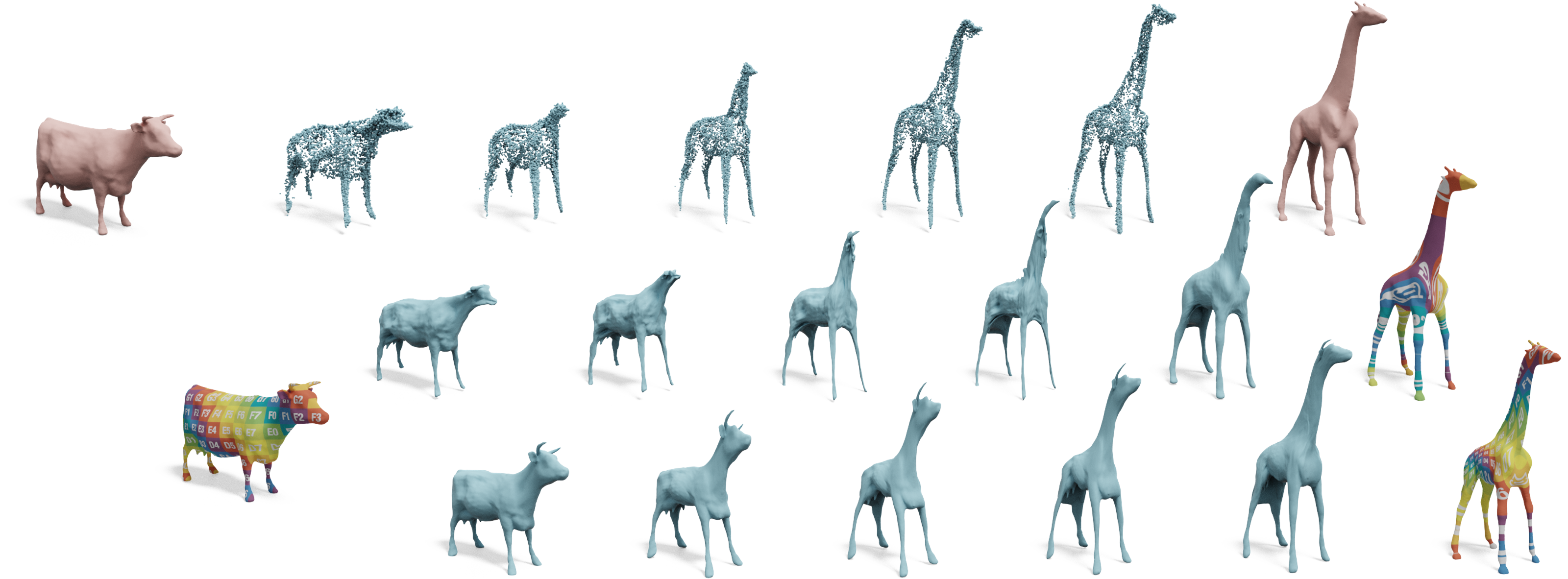}
  \end{center}
\caption{We select a challenging pair for registration result visualization. The first row shows the point clouds we obtain for guidance, which contain significant noise due to the large deformations undergoing between the shapes. The second row demonstrates the results of using a naive iterative registration approach, which exhibits poor robustness to the noise in the point clouds. The third row displays our results using the continuous normalizing flow, achieving high-quality registration and reasonable intermediate interpolation results.}
\label{fig:interpolation3row}
\end{figure*}

Though it seems natural to iteratively perform shape registration~\cite{nicp} between consecutive point clouds in $\mathcal{V}$, we observe that this naive approach can lead to sub-optimal results. As shown in the top row of Fig.~\ref{fig:interpolation3row}, due to the significant deformation between the cow and the giraffe, the intermediate point clouds are of low quality. Deforming the cow towards such can lead to distortion accumulation (see the middle row of Fig.~\ref{fig:interpolation3row}). 

In light of this, we propose a more global consistent registration scheme. Namely, we cast the registration problem as estimating a \emph{flow} that deforms $\src$ towards $\tar$, while approximating the intermediate point clouds at the regarding time steps. 
As shown in the bottom row of Fig.~\ref{fig:interpolation3row}, we achieve high-quality semantic correspondences in this challenging case, while being robust regarding imperfect intermediate guidance. 

In particular, we let $y(t)$ be a continuously deforming point cloud with respect to temporal parameter $t$, such that $y(t_0) = V_0$, our target is then to learn a continuous-time dynamic $\frac{\partial y(t)}{\partial t}=f(y(t),t)$, or, a flow,  that indicates how $y(t)$ evolves over time.

In order to estimate the flow, we adopt the framework of PointFlow~\cite{yang2019pointflow}. 
The key motivations are: 1) to achieve an invertible normalizing flow; 2) to exploit the powerful MLP-based flow representation. 
In~\cite{yang2019pointflow}, $f$ is represented by a neural network with an unrestricted architectural design. 
A Continuous Normalizing Flow (CNF) models an entity $x$ with an initial prior distribution at the starting time as $x=y(t_0)+\int_{t_0}^{t_1}f(y(t),t)dt,\quad y(t_0)\thicksim P(y)$. The value at $y(t_0)$ can be obtained using the inverse of the flow, expressed as 
\begin{equation}\label{eqn:1}
y\left(t_0\right)=x+\int_{t_1}^{t_0} f(y(t), t) dt.
\end{equation}

In our context, we do not assume prior distributions for $x$ and $y$ but rather treat it as a registration problem. 
Here we set $x$ as the source point cloud $\src$, and $y(t_0)$ as a prediction of the target point cloud $\tar$. 
We define $f(y(t),t)$ as a multi-layer perceptron (MLP). 
Ultimately, we can obtain the predicted value of the target by solving the ODE (Ordinary Differential Equation) from Eqn.~\ref{eqn:1}.

For fitting the flow, we utilize the following energy function. 
\begin{equation}\label{eqn:2}
E_{\mbox{cd\_final}} = CD(y(t_0), V_{T+1}),
\end{equation}
where CD refers to Chamfer distance, which is widely used to measure the extrinsic distance between two point clouds.
To reduce the complexity of optimizing the aforementioned MLP, we sample $k$ discrete time points on the continuous flow as $y(t_k)$. 
To provide additional guidance throughout the process, we assign functions at specific time points based on the point cloud $\mathcal{V} = \{V_{t} \mid t = 1,...,T \}$ obtained from the previous section. 
The specific cost function $E_{\mbox{cd\_inter}}$ is. 
\begin{equation}\label{eqn:3}
E_{\mbox{cd\_inter}} = \frac{1}{T} \sum_{ i \in [T] } CD(y(t_i), V_{i}).
\end{equation}
In addition, to ensure the shape does not undergo excessive distortion during the registration process, we include an ARAP (As-Rigid-As-Possible) loss followed by~\cite{guo2021human} as a regularization term. 
The total cost function $E_{\text {total }}$ combines the above terms with the weighting factors $\lambda_{\mbox{cd\_inter}}, \lambda_{\mbox{cd\_final}}$, and $\lambda_{\mbox{arap}}$ to balance them:
\begin{equation}\label{eqn:total}
E_{\mbox{total}}=\lambda_{\mbox{cd\_inter}} E_{\mbox{cd\_inter}}+\lambda_{\mbox{cd\_final}} E_{{\mbox{cd\_final}}}+\lambda_{\mbox{arap}} E_{\mbox{arap}}.
\end{equation}

To achieve more accurate registration results, we perform non-rigid shape registration~\cite{nicp} between the output of the flow network and $\tar$.

\section{Experimental Results}\label{sec:exp}

\subsection{Implementation Details}\label{sec:implement}

For the input source and target shape, we use Open3D to sample uniformly $16$ viewpoints. 
In particular, empirically we render the background of all rendered images to black for the best overall performance. 
Note that we demonstrate a white background in Fig.~\ref{fig:pipeline}, which is for better visualization. 
We use DiffMorpher~\cite{diffmorpher} to interpolate $10$ frames between each pair of images rendered from the same viewpoint. 
Regarding SC-GS~\cite{sc-gs}, we use vertices extracted from the source mesh as the initialization of the mean of Gaussians. 
The first $3000$ iterations of the training step are used for initializing each single Gaussian. 
Then we train a model to predict the deformation between them, simultaneously optimizing the position, opacity, and covariance matrix for a total of $20000$ iterations. We set the percent dense parameter to $0.01$ to generate relatively sparse point clouds. In general, we believe in target shape more than the intermediate point clouds. During flow estimation, we set the weight $\lambda_{\mbox{cd\_inter}}$ to $1$, that of $\lambda_{\mbox{cd\_final}}$ to $10$, and $\lambda_{\mbox{arap}}$ to $2$ for all categories. We train $4000$ iterations per pair and the learning rate is set to $1e-3$. We set $t_0 = 0$, $t_1 = 0.5$ and using the $2nd, 4th, 6th$ and $8th$ of the $T = 10$ reconstructed point clouds in Eqn.~\ref{eqn:3}.

\begin{table*}[ht]
\caption{Average geodesic errors on each of the test categories in SHREC07~\cite{shrec07} and EBCM~\cite{EBCR}. 
}\label{table:geo_error}
\centering
\begin{tabular}{ccccccccccc}
\hline
\rowcolor[HTML]{FFFFFF} 
            & Airplane        & Ant             & Bird            & Chair           & Fish            & Fourleg         & Glasses         & Human           & Plier           & EBCM                         \\ \hline
\rowcolor[HTML]{FFFFFF} 
MapTree     & 0.5634          & 0.2635          & 0.4117          & 0.4742          & 0.2949          & 0.3507          & 0.6878          & 0.2633          & 0.2805          & 0.4231                    \\
\rowcolor[HTML]{FFFFFF} 
BIM         & 0.2589          & 0.3050          & 0.4123          & 0.4800          & 0.2699          & 0.2449          & 0.6160          & 0.1810          & 0.5197          & 0.2968                    \\
\rowcolor[HTML]{FFFFFF} 
SmoothShell & 0.2749          & 0.2694          & 0.3009          & 0.2627          & 0.1032          & 0.1234          & 0.3100          & 0.1572          & 0.3900          & 0.4476                    \\
\rowcolor[HTML]{FFFFFF} 
NeuroMorph  & 0.1510          & 0.1895          & 0.1672          & 0.1853          & 0.1366          & 0.1697          & 0.2652          & 0.2141          & 0.2830          & 0.1844                    \\
\rowcolor[HTML]{FFFFFF} 
ENIGMA      & 0.3568          & 0.3675          & 0.3278          & 0.4482          & 0.1733          & 0.2466          & 0.6187          & 0.2925          & 0.4379          & 0.3032                    \\
\rowcolor[HTML]{E7E6E6} 
Ours        & \textbf{0.0356} & \textbf{0.1133} & \textbf{0.0965} & \textbf{0.0295} & \textbf{0.0804} & \textbf{0.0824} & \textbf{0.0614} & \textbf{0.1467} & \textbf{0.1476} & \textbf{0.0886} \\ \cline{1-11}
\end{tabular}
\end{table*}
\subsection{Evaluation Setup}~\label{sec:eval}

\noindent\textbf{Baselines:} In this section, we compare our method with an array of methods of estimating dense correspondences, which 1) require no external landmarks as input and 2) pose no constraint (\emph{e.g., } near-isometry) on the deformations between input shapes: BIM~\cite{kim2011blended}, SmoothShells~\cite{smoothshells}, NeuroMorph~\cite{eisenberger2021neuromorph}, MapTree~\cite{maptree}, and ENIGMA~\cite{enigma}

\noindent\textbf{Benchmarks:} We comprehensively compare our method and the baselines in an array of test sets. (1) We consider $9$ categories from SHREC07 dataset~\cite{shrec07} -- \texttt{human, fourleg, airplane, bird, chair, fish, ant, pier}, and \texttt{glasses}, each of which contains $20$ shapes. 
In order to fully evaluate the capacity of all methods, we select the most distinctive shape pairs via the following scheme. 
For a pair of shapes $S_i, S_j$ from the same category, we compute the first $50$ eigenvalues $\Lambda_i, \Lambda_j$ and the spectral distance $d(S_i, S_j) = \Vert \Lambda_i - \Lambda_j\Vert$. 
We then sample $50$ pairs from the overall $190$ in the descending order of spectral distances for our test. 
(2) We further pick ten pairs from the test cases presented in~\cite{EBCR} for more variability. 
Finally, we remark that for each pair $(S_i, S_j)$, we compute maps in both directions.

\noindent\textbf{Evaluation Metrics:} 
To evaluate the maps, we consider four well-known metrics including Dirichlet energy~\cite{rhm}, Coverage (\emph{i.e.} surjectivity)~\cite{huang2017adjoint}, Geodesic matching error with respect to ground truth landmark labels~\cite{kim2011blended}, and Bijectivity~\cite{BCICP}. 
We refer readers to the Supp. Mat. for the details on these metrics. 
\begin{figure}[t!]
  \begin{center}
\includegraphics[width=7cm]{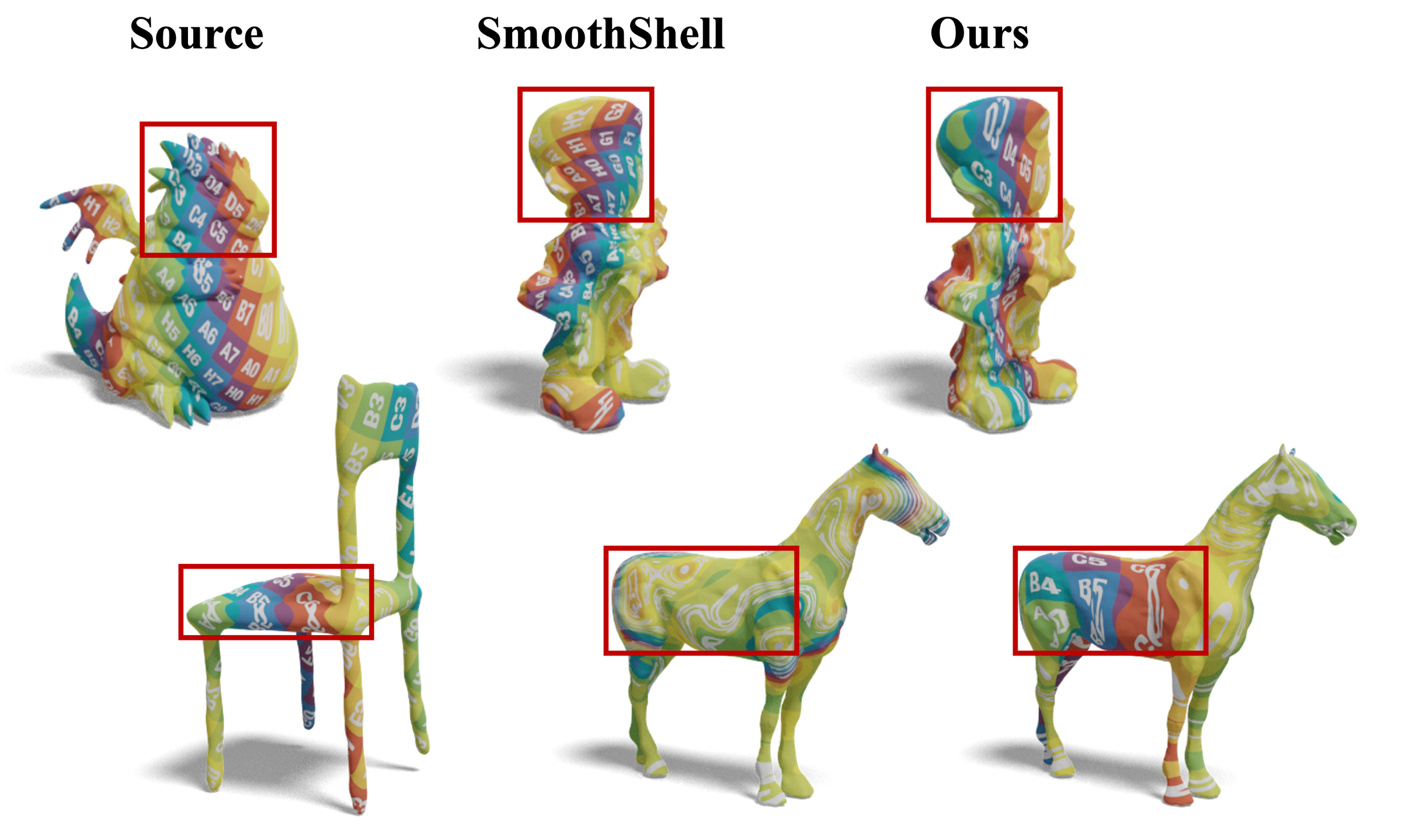}
  \end{center}
  \vspace{-3mm}
\caption{We match shapes from distinctive categories. Top: Dragon vs. Solider; Bottom: Chair vs. Horse. }
\label{fig:crossclass}
\end{figure}

\subsection{3D Shapes Interpolation}\label{sec:app}

\begin{figure*}[h!]
  \begin{center}
\includegraphics[width=17.5cm]{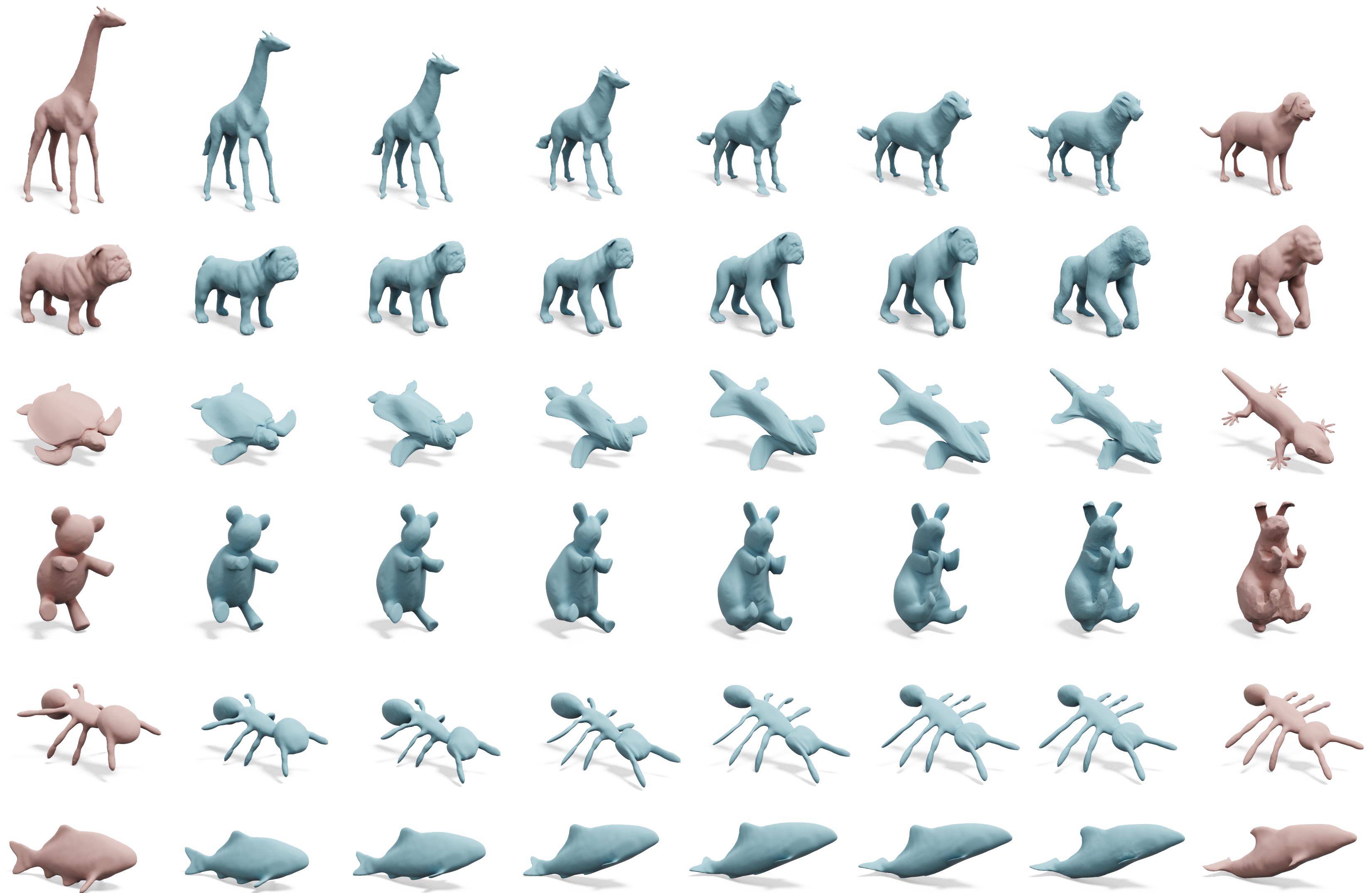}
  \end{center}
\caption{Demonstrations of smooth and semantically meaningful shape interpolations obtained by our estimated flow. }
\label{fig:interpolation}
\end{figure*}

In Fig.~\ref{fig:interpolation}, we visualize interpolations between shapes induced by our learned flow, which are smooth, plausible, and semantically meaningful. 
Beyond achieving high-quality maps, we believe that they also reveal the potential of our method in generating high-quality 3D assets autonomously.

\begin{figure}[t!]
  \begin{center}
\includegraphics[width=7cm]{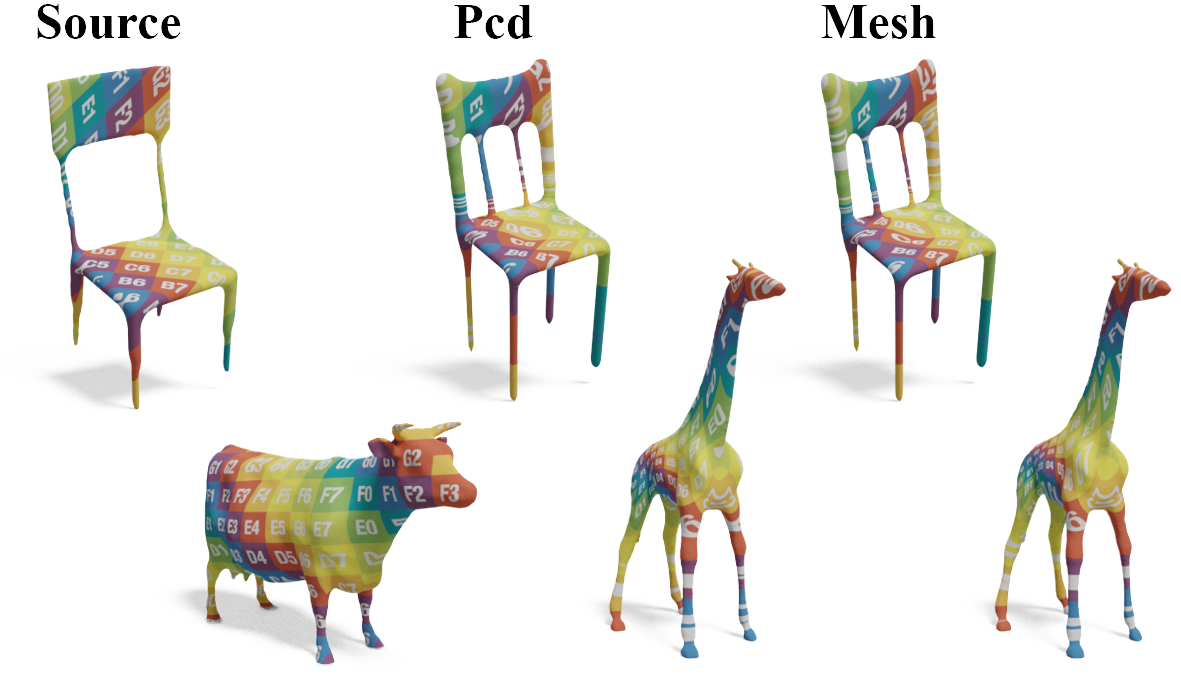}
  \end{center}
 \vspace{-4mm}
\caption{Our framework can be directly applied to unstructured point clouds. We visualize maps computed with point cloud input (middle) and mesh input (right). Both maps are comparable and semantically meaningful. }
\label{fig:pcd_mesh}
\end{figure}

\subsection{Dense Shape Correspondence}\label{sec:dense}
We report both quantitative and qualitative results on the involved benchmarks. 
First of all, we report the average geodesic errors of each test set in Tab.~\ref{table:geo_error}.
Our method outperforms \emph{all} baselines across \emph{all} sets, with significant margins in the categories of \texttt{airplane, ant, bird, chair, glasses, piler} and \texttt{EBCM}. 

In total, \textbf{SRIF} achieves an average Geodesic Error of $0.0956$, which is less than half of the second-best baseline.
Our method even outperforms ENIGMA in Dirichlet Energy -- our method does not explicitly optimize for this metric while ENIGMA uses RHM~\cite{rhm} for post-processing, which minimizes Dirichlet energy. 
Fig.~\ref{fig:shrec07} showcases part of the qualitative results, which agree nicely with the quantitative results. 
One obvious problem among the intrinsic-based method (BIM, MapTree, Enigma) is symmetry ambiguity. 
Though our method gets rid of such by extrinsic alignment, we argue that such a requirement is indeed easier to meet than injecting orientation information into intrinsic-based methods. 
Moreover, in the Supp. Mat., we also report the scores of ENIGMA with allowance on the symmetric flip (as in the regarding paper), yet our method still outperforms it across all sets with more restrictive evaluation.

We further consider cross-category pairs in Fig.~\ref{fig:crossclass}. 
Though both methods produce distorted maps in the two challenging pairs, our method better captures the structural correspondences between shapes (see the red boxes).

\begin{table}[!t]
\caption{Average scores regarding over all categories in Tab.~\ref{table:geo_error}.
}\label{table:results}
\centering
\begin{tabular}
{
>{\columncolor[HTML]{FFFFFF}}c 
>{\columncolor[HTML]{FFFFFF}}c 
>{\columncolor[HTML]{FFFFFF}}c 
>{\columncolor[HTML]{FFFFFF}}c 
>{\columncolor[HTML]{FFFFFF}}c }
\hline
             &  Dirichlet ↓ & Cov. ↑  & Lmks. Err. ↓ & Bij ↓                   \\ \hline

MapTree     & 17.7309                                                         & 0.3967                  & 0.3683                                                           & 0.0432                  \\
BIM         & 12.4723                                                         & 0.4665                  & 0.3200                                                           & 0.2504                  \\
SmoothShells & 14.0198                                                         & 0.6275                  & 0.2221                                                           & 0.0101                  \\
NeuroMorph  & 22.0461                                                         & 0.1099                  & 0.1931                                                           & 0.0944                  \\
ENIGMA     & 6.5344                                                 & 0.6168                  & 0.3464                                                           & 0.0123                 \\
\rowcolor[HTML]{E7E6E6} 
Ours        & \textbf{6.4702}                                                 & \textbf{0.6418} & \textbf{0.0956}                                          & \textbf{0.0075} \\ \hline
\end{tabular}
\end{table}

\subsection{Point-based \textbf{SRIF}}\label{sec:pbsrif}
In fact, \textbf{SRIF} can be directly applied on \emph{unstructured point clouds}. 
The only two parts in Sec.~\ref{sec:method} where we utilize mesh information are multi-view rendering and the construction of local neighborhoods for ARAP regularization. 
For the former, the Open3D library supports rendering point clouds, despite of a certain degree of detail loss; For the latter, one can approximate Euclidean proximity among vertices. As shown in Fig.~\ref{fig:pcd_mesh}, our framework manages to deliver high-quality correspondences with less structured input. 
This is in sharp contrast to methods heavily depending on surface geometry, for instance, the spectral-based shape matching techniques.

\subsection{Robustness}\label{sec:add}
We demonstrate the our robustness in the following perspectives: 

\begin{figure}[t!]
  \begin{center}
\includegraphics[width=8.5cm]{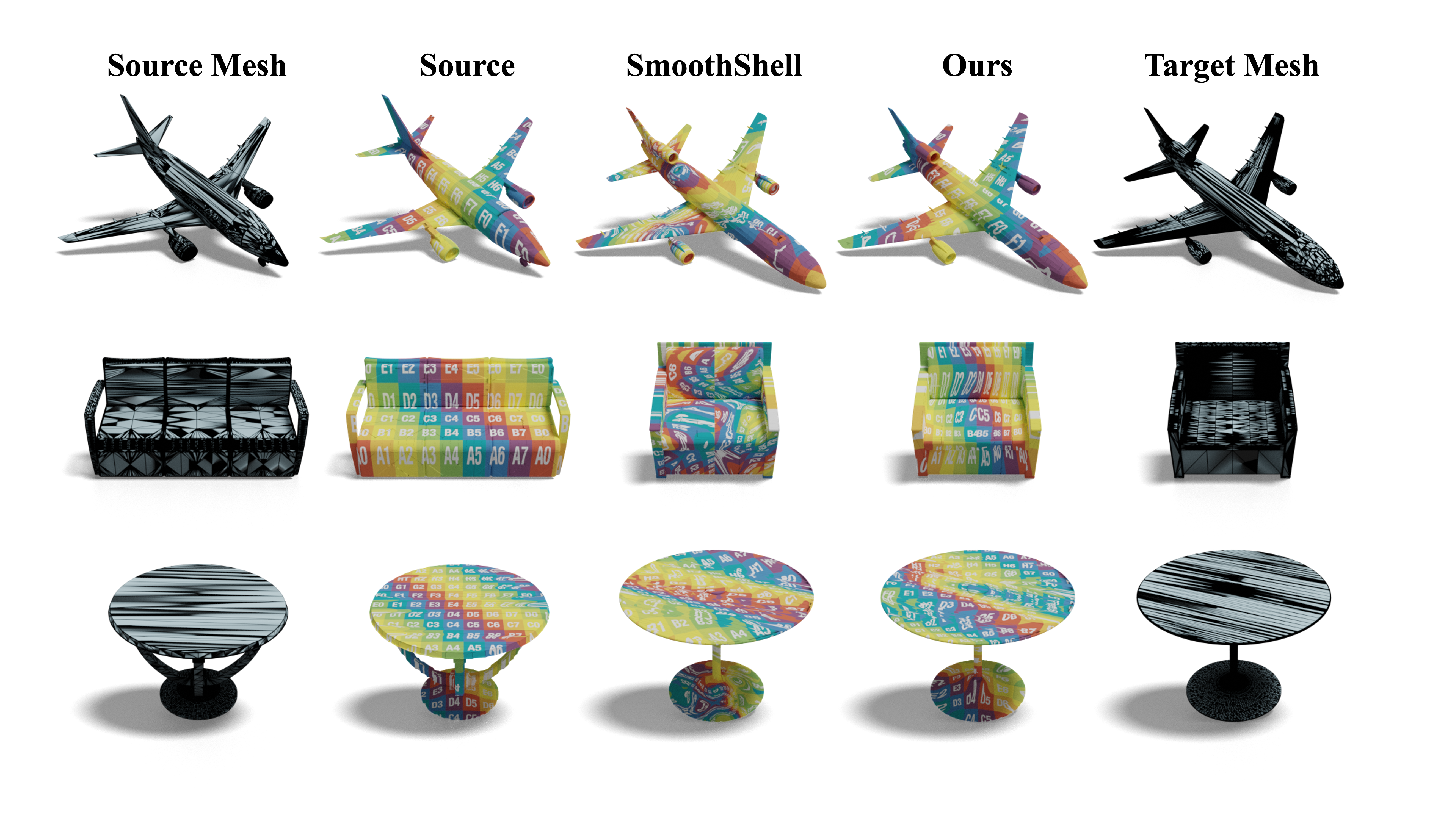}
  \end{center}
    \vspace{-7mm}
\caption{In practice, meshes can exhibit server irregularities (left-most and right-most columns). Our method demonstrates clearly better robustness than SmoothShells (top and middle rows), while can fail in the presence of large topological changes (bottom row). }
\label{fig:keypointnet}
\end{figure}
\noindent\textbf{Mesh Quality: }In practice, meshes can exhibit severe irregularities. For instance, the shapes in the right-most and left-most columns of Fig.~\ref{fig:keypointnet} are from KeyPointNet dataset~\cite{keypointnet}, which consist of a large portion of thin triangles. 
This can pose significant challenges on registration methods based on triangulation and point-wise correspondence update. 
For instance, in the top row of Fig.~\ref{fig:keypointnet}, SmoothShells cannot match the airplanes even though they are similar to each other.
In contrast, thanks to the flow estimation module, our method enjoys stronger robustness on this front.

\begin{table}[!t]
\caption{We validate the robustness of our method to rotation on one pair by rotating the target around the X, Y, Z axes by 10, 30, and 45 degrees, respectively, and calculating the landmark errors.
}\label{table:rot}
\centering
\begin{tabular}{
>{\columncolor[HTML]{FFFFFF}}c 
>{\columncolor[HTML]{FFFFFF}}c 
>{\columncolor[HTML]{FFFFFF}}c 
>{\columncolor[HTML]{FFFFFF}}c 
>{\columncolor[HTML]{FFFFFF}}c }
\hline
  & 0                                                         & 10     & 30     & 45     \\ \hline
X & \cellcolor[HTML]{FFFFFF}                                  & 0.0529 & 0.0535 & 0.0549 \\
Y & \cellcolor[HTML]{FFFFFF}                                  & 0.0555 & 0.0563 & 0.0565 \\
Z & \multirow{-3}{*}{\cellcolor[HTML]{FFFFFF}\textbf{0.0529}} & 0.0549 & 0.0551 & 0.0657 \\ \hline
\end{tabular}
\end{table}
\noindent\textbf{Rotational Perturbation: }We generally assume the shapes of interest are extrinsically aligned, which is a common practice in shape registration~\cite{AMM}. 
Meanwhile, we empirically observe that \textbf{SRIF} is robust with rotational perturbation as large as $45$ degrees (see Tab.~\ref{table:rot} and qualitative results in the Supp. Mat.). 
We attribute such to the fact that the diffusion model is trained with objects in various orientations~\cite{3dprobing}, thus gaining certain robustness and passing to our framework.

\subsection{Ablation Studies}\label{sec:abl}

\noindent\textbf{Intermediate geometry:} 
It is worth noting that our flow estimation can be conducted with as few as two shapes. 
We report the direct estimation in the first row of Tab.~\ref{table:abl}, which clearly suggests the necessity of applying LVM to achieve the intermediate point clouds. At the same time, we ablate the surface point cloud extraction and use the original Gaussian for guidance, whose absence causes performance degradation.

\noindent\textbf{Reconstruction Method:}
We perform dynamic 3D-GS on the whole set of multi-view interpolated images. 
One can as well independently perform 3D Gaussian Splatting~\cite{3dgs} on the multi-view images at each time step to obtain the intermediate geometry. 
The third row of Tab.~\ref{table:abl} suggests that the above variant is suboptimal. This is probably since image interpolation is performed \emph{independently} from each viewpoint, therefore it is hard to guarantee the multi-view consistency at each time step.

\noindent\textbf{View Number }is an important hyper-parameter of our method, in the fourth and fifth row of Tab.~\ref{table:abl}, we ablate it by testing with fewer views. 
It is evident that a larger number of views is advantageous, as it naturally covers more thoroughly the shapes of interest. 
Of course, as will be discussed in Sec.~\ref{sec:time}, increasing the view number would significantly slow down the method, we choose $16$ with a trade-off of efficiency and accuracy. 

\begin{table}[!t]
\caption{Comparison to several variants of our method on the pair shown in Fig.~\ref{fig:landmark}, see the main text for details. 
}\label{table:abl}
\centering
\begin{tabular}
{
>{\columncolor[HTML]{FFFFFF}}c 
>{\columncolor[HTML]{FFFFFF}}c 
>{\columncolor[HTML]{FFFFFF}}c 
>{\columncolor[HTML]{FFFFFF}}c 
>{\columncolor[HTML]{FFFFFF}}c 
>{\columncolor[HTML]{FFFFFF}}c }
\hline
             & Dirichlet ↓ & Cov. ↑                      & Lmks. Err. ↓        & Bij ↓             \\ \hline
Direct flow est. & 16.1538 & 0.2881                & 0.1157 & 0.0463          \\
w/o surface ext. & 6.9506 & 0.6988                & 0.0665 & 0.0069          \\
w/ 3D-GS      & 8.9399           & 0.5536                     & 0.0833          & 0.0159          \\
4 views        & 7.7054           & 0.6917              & 0.0764          & 0.0072          \\
8 views        & 6.6097           & 0.6963               & 0.0663          & 0.007           \\
Full         & \textbf{6.3094}  & \textbf{0.7031} & \textbf{0.0529} & \textbf{0.0064} \\ \hline
\end{tabular}
\end{table}

\begin{figure}[t!]
  \begin{center}
\includegraphics[width=7cm]{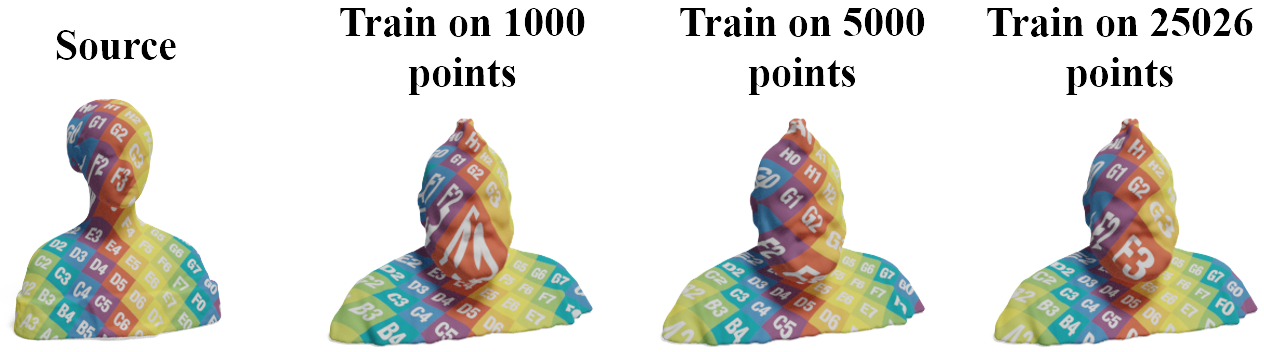}
  \end{center}
  \vspace{-4mm}
\caption{We test the robustness of our flow estimation to the number of input points. We train it on 1,000 points, 5,000 points, and the original 25,026 points, respectively, and conduct the tests on the original point cloud.}
\label{fig:scalability}
\end{figure}

\subsection{Running Time Analysis}\label{sec:time}
We compare time efficiency of our method and baselines with respect to a fixed pair of shapes on the same machine (see the Supp. Mat. for details). MapTree, BIM, NeuroMorph all take 1 min, and SmoothShells takes 5 mins.
ENIGMA is ran by the authors, who report running time of 20 mins to process shapes of 5000 vertices with post-refinement. 
Our method takes 40 mins (20 mins for image morphing, 10 mins for intermediate point clouds generation, and 10 mins for flow estimation). 
The complexity of image morphing and 3D Gaussian reconstruction is determined by the number of views and interpolating images. 
The complexity of flow estimation depends on the number of vertices on $\src$. 

While efficiency remains the main bottleneck, we highlight that, compared to the baselines, our method achieves significantly more precise maps but also high-quality morphing process across various categories. 
Furthermore, since our flow estimation module can learn a continuous flow with finite discrete point clouds, we can downsample $\src$ and follow the strategy in Sec.~\ref{sec:pbsrif} to alleviate the computational burden. 
As shown in Fig.~\ref{fig:scalability}, significant down-sampling leads to a reasonable performance drop. 
We finally emphasize that the flow trained on the down-sampled point cloud can be inferred on the original one directly, without any post-processing.

\begin{figure*}[h!]
  \begin{center}
\includegraphics[width=17.5cm]{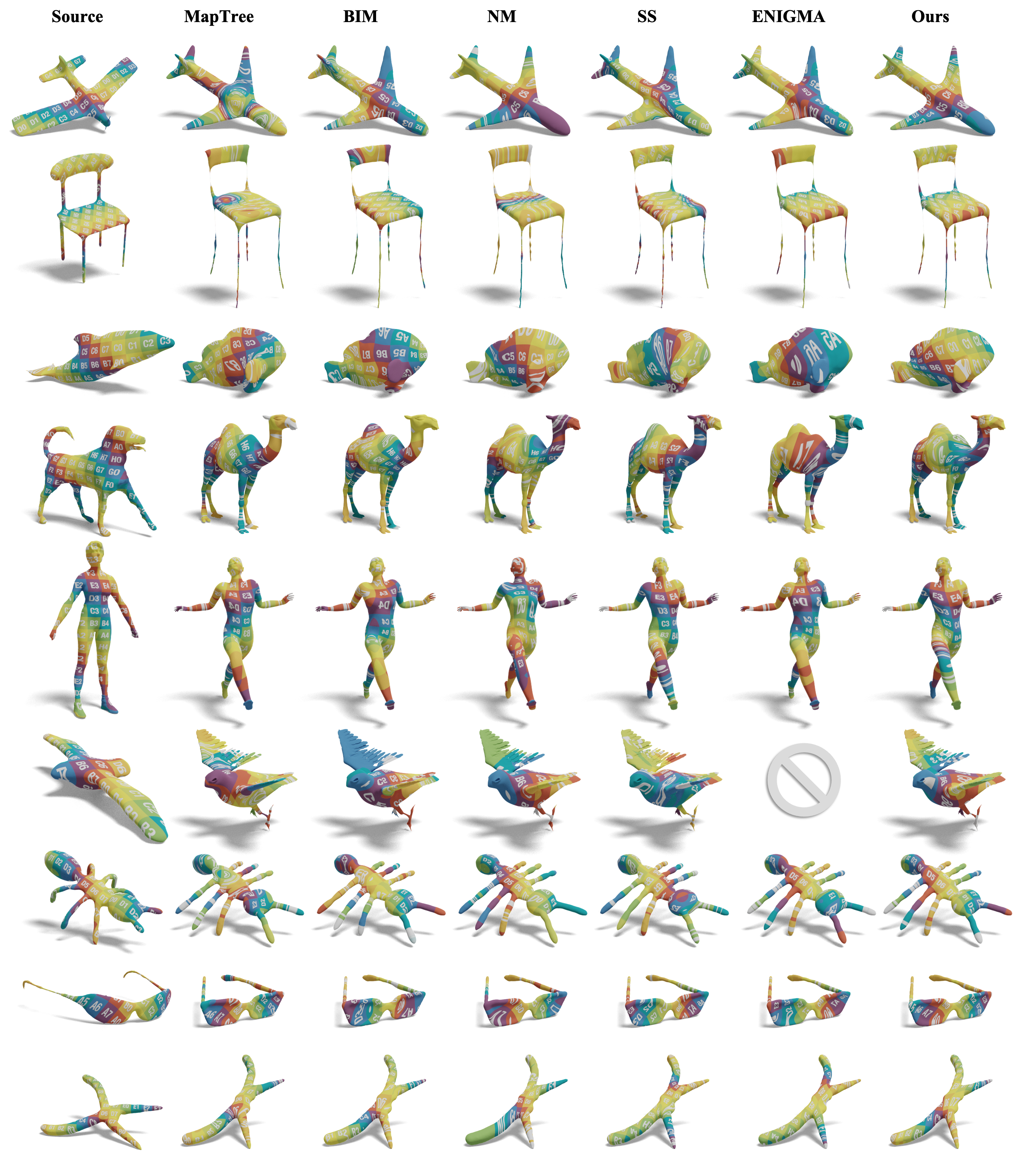}
  \end{center}
\caption{Qualitative comparisons on various categories from SHREC'07~\cite{shrec07}. ENIGMA~\cite{enigma} fails to return result on the pair of birds. Note that the target chair in the second row is \emph{not} disconnected -- It appears so due to the thin structure within the chair. }
\label{fig:shrec07}
\end{figure*}

\section{Conclusion and Limitations}
\begin{table}[!b]
\caption{Mean geodesic errors on the SMPL and SHREC19 dataset.}\label{table:smpl}
\centering
\begin{tabular}{
>{\columncolor[HTML]{FFFFFF}}c 
>{\columncolor[HTML]{FFFFFF}}c 
>{\columncolor[HTML]{FFFFFF}}c 
>{\columncolor[HTML]{FFFFFF}}c 
>{\columncolor[HTML]{FFFFFF}}c 
>{\columncolor[HTML]{FFFFFF}}c }
\hline
        & Maptree & BIM & SS & NM & Ours   \\ \hline
SMPL    & 0.1715  &  0.1329   & 0.0354      & 0.1017     & 0.0696 \\
SHREC19 & 0.3013  &  0.2174  & 0.0685      & 0.1499     & 0.0823 \\ \hline
\end{tabular}
 \vspace{-1em}
\end{table}
In this paper, we propose \textbf{SRIF}, an autonomous framework for semantic 3D shape registration. 
By exploiting semantic information obtained from LVMs in a dynamic manner and with a novel flow estimation module, \textbf{SRIF} achieves high-quality dense correspondences on challenging shape pairs, but also delivers smooth, semantically meaningful interpolation as a by-product. 
Ablation studies justify our overall design and highlight the robustness and scalability of our framework. 

We identify the following limitations of our method, which lead to future work directions: 
1) There exists significant room for improvement on efficiency. 
As shown in Sec.~\ref{sec:time}, the main bottleneck of our pipeline lies in image morphing, or, more specifically, LoRA fine-tuning, which takes over $50\%$ of the total running time. 
To this end, we plan to resort to advances in parameter efficient fine-tuning; 
2) Our method does not guarantee the output to be continuous or bijective. 
It would be interesting to explore how to regularize the smoothness of flow~\cite{vp}; on the other hand, since flow is by construction invertible, we can encourage bijectivity by taking forward and backward flow simultaneously during training; 
3) Since our method leverages image interpolation, it could be suboptimal when the intermediate results are problematic. 
To see that, we evaluate our method on SHREC19~\cite{SHREC19} and SMPL~\cite{SMPL} dataset. 
As shown in Tab.~\ref{table:smpl}, our method is outperformed by SmoothShells, which leverages intrinsic geometry information, with a notable margin.  
We attribute the failure to two factors: first human articulation spans a large space, which might not be well learnt by diffusion models without explicit modeling; second there exist self-occlusions among the multi-view renderings of human shapes. 
We refer readers to the Supp. Mat. for more details on this experiment. 
We plan to introduce stronger prior on this front, say, leveraging pre-trained model on human shapes/images. 
4) Our method uses ARAP to regularize mesh deformation, which implicitly encourages local neighborhood preservation. This in turn prevents our method from characterizing significant topological changes (see bottom row of Fig.~\ref{fig:keypointnet}, where the supports of tables are distinctive). 
As shown in Sec.~\ref{sec:pbsrif}, our method can be adapted into purely point-based settings, it is interesting to further exploit this property.

\noindent\paragraph*{\textbf{Acknowledgement}} This work was supported by the National Natural Science Foundation of China under contract No. 62171256 and Meituan.

\appendix
\section{Appendix}
In this supplementary material, we provide ablation on our color scheme of choice (Sec.~\ref{supp:color}); more details about the surface point extraction module (Sec.~\ref{supp:surf}); detailed formulations on the metrics (Sec.~\ref{supp:metric}); qualitative results on robustness regarding misalignment (Sec.~\ref{supp:rot}); qualitative comparison to landmark-based approach (Sec.~\ref{supp:smat}); details on the non-rigid human shape matching in Sec. 5 of the main paper (Sec.~\ref{supp:human}); details on running time analysis (Sec.~\ref{supp:time}); per-category quantitative results regarding Tab.1 in the main paper (Sec.~\ref{supp:percate}). 

\subsection{Ablation study on color scheme for rendering}\label{supp:color}

As we put no assumption on the texture of 3D shapes of interest, the rendered images often suffer from loss of details. 
On the other hand, diffusion models are trained on realistic images with rich texture. 
To compensate the discrepancy, we design a specific color coding scheme to add more texture details.

\begin{figure}[t!]
  \begin{center}
\includegraphics[width=8cm]{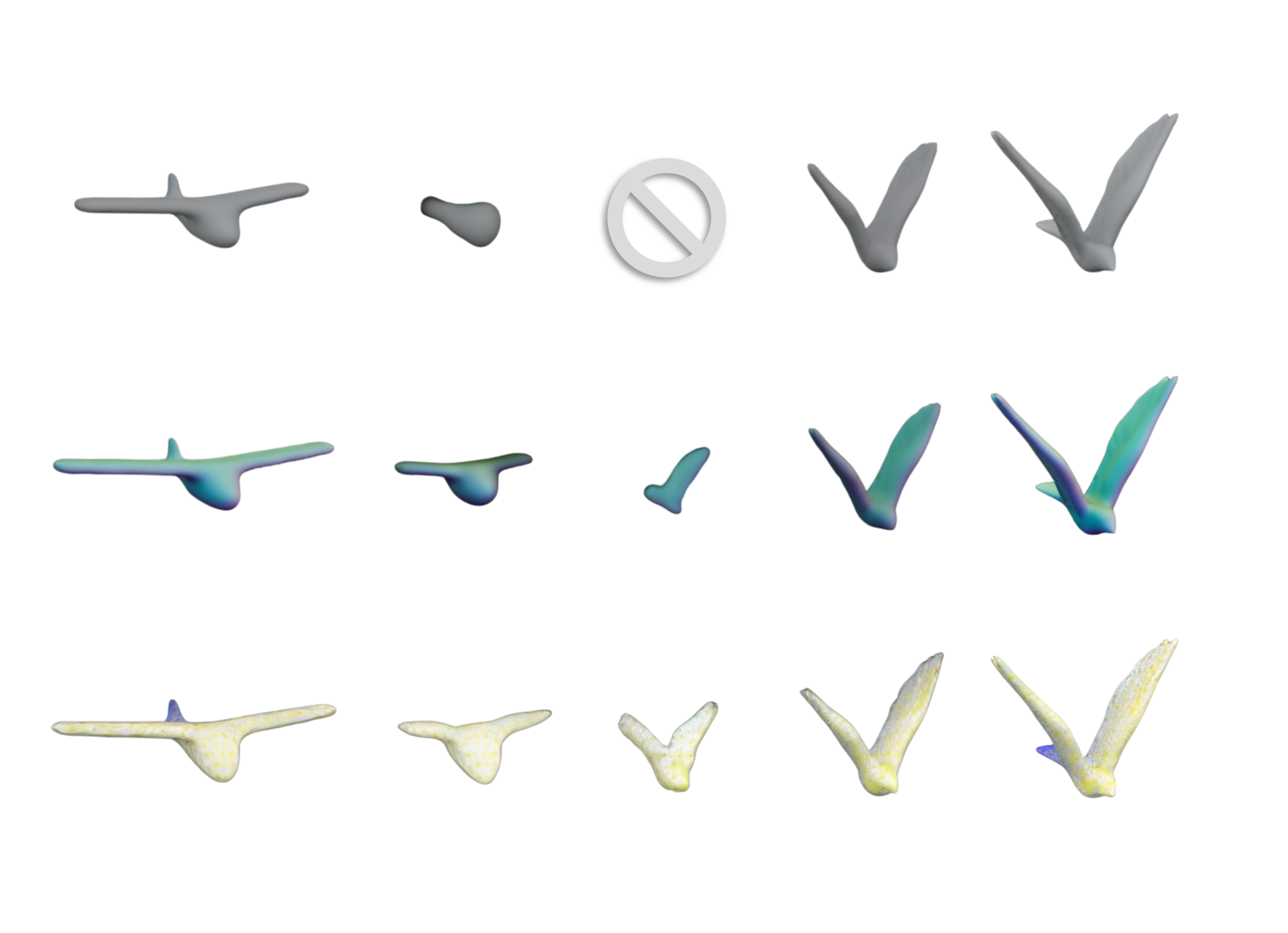}
  \end{center}
\caption{We use 3 different color schemes to render the mesh. The first row shows rendering without textures, and the object disappears in the middle frame. The second row shows rendering with normal vector coloring, and the right wing of the target bird is still close to disappearing. The third row shows our coloring method, where the interpolation sequence is smooth and plausible.}
\label{fig:texture}
\end{figure}

In this part, we compare the effect of three different color schemes, including textureless rendering, normal-based color scheme, and the one proposed in Sec.3.1 of the main paper. 
As shown in Fig~\ref{fig:texture}, our color scheme yields the most natural and complete interpolation between an airplane and a bird.
The rest two, on the other hand, suffer from either missing frames or missing parts.

\subsection{Details on surface point extraction module}\label{supp:surf}

\begin{figure}[h!]
  \begin{center}
\includegraphics[width=6.5cm]{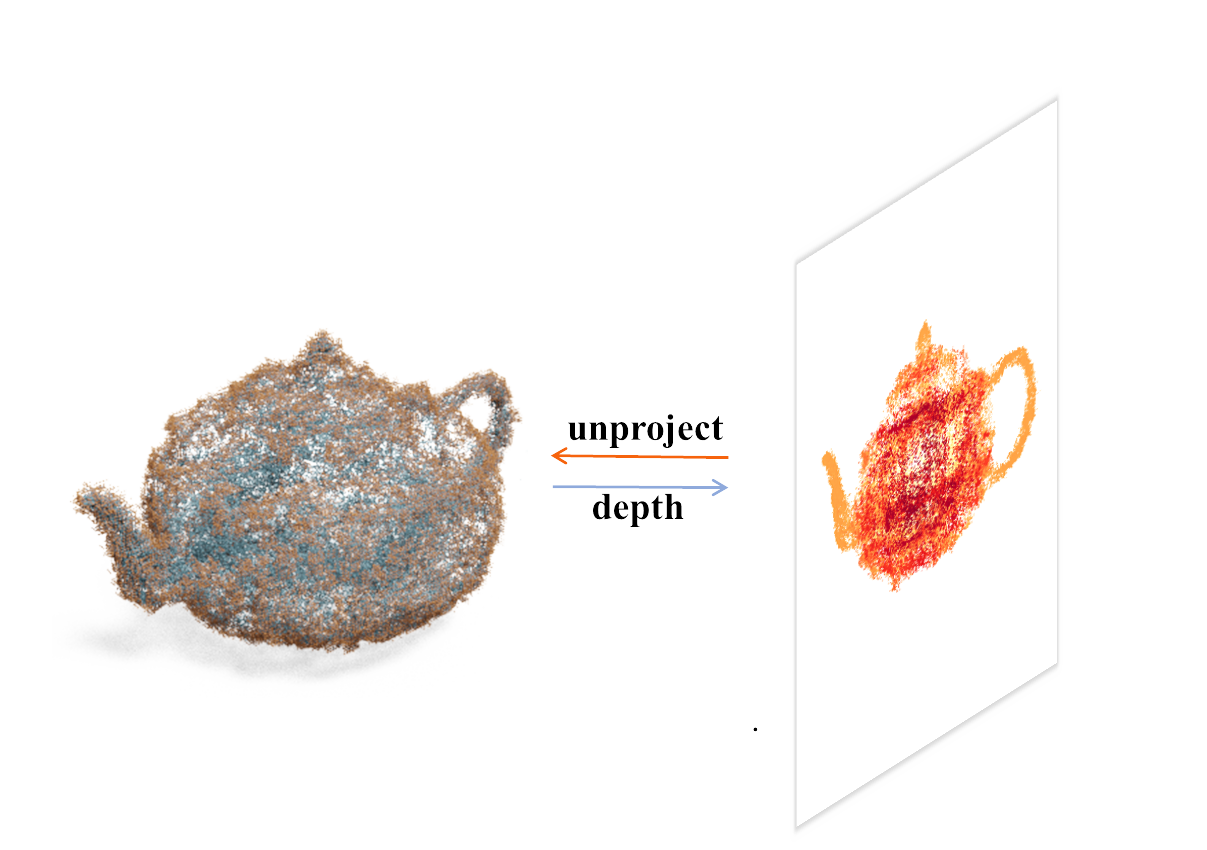}
  \end{center}
\caption{Our surface point extraction module operates by projecting and unprojecting depth images from multiple viewpoints. To better illustrate our method, we only show one viewpoint in the figure.}
\label{fig:surface}
\end{figure}

Figure~\ref{fig:surface} shows our surface points extraction operation. 
We assume to be given an input point cloud $\mathcal{P}$ as well as a set of camera positions $\mathcal{E} = \{\mathbf{e}_1, \mathbf{e}_2, \ldots, \mathbf{e}_N\}$  distributed on a sphere surrounding the point cloud. 
For each camera position $\mathbf{e}_i$, the camera is configured with parameters including the field of view $\theta$, center point $\mathbf{c}$, and up vector $\mathbf{u}$. 
A depth image $\mathcal{D}_i$ is rendered from the perspective of the current camera position. 
The rendered depth image $\mathcal{D}_i$ is then unprojected to obtain the corresponding 3D points $\mathcal{Q}_i$ in the world coordinate system.
\begin{align}
    \mathbf{q} &= \mathcal{U}(\mathbf{p}, \mathcal{D}i(x, y), w, h),\
\end{align}
where $\mathbf{q}$ is the unprojected 3D point, $\mathbf{p} = (x, y)$ is a pixel in the depth image, and $w$ and $h$ are the width and height of the depth image, respectively.

Intuitively, the inner points are filtered out through a combination of forward depth image rendering and inverse unprojection. 
In other words, only points around the surface are extracted.

\subsection{Evaluation Metrics}\label{supp:metric}
We use the following evaluation metrics to assess the quality of the generated maps and registration results.

\noindent\textbf{Dirichlet Energy:} The Dirichlet energy measures the smoothness of the mapping between the source and target shape. It is defined as:
\begin{equation}
        E_D(f) = \frac{1}{2} \int_\mathcal{S} |\nabla f|^2 dA,
\end{equation}
where $f: \mathcal{S} \rightarrow \mathcal{T}$ is the mapping between the source shape $\mathcal{S}$ and the target shape $\mathcal{T}$, and $\nabla f$ denotes the gradient of $f$. 
A lower Dirichlet energy indicates a smoother mapping.

\noindent\textbf{Coverage (surjectivity):} Coverage evaluates the extent to which the target shape is covered by the mapped image of the source shape. It is defined as:
    \begin{equation}
        \text{Coverage}(f) = \frac{|{q \in \mathcal{T} : \exists p \in \mathcal{S}, f(p) = q}|}{|\mathcal{T}|},
    \end{equation}
    where $|\cdot|$ denotes the cardinality of a set. 
    A higher Coverage score suggests a more injective mapping.

    \noindent\textbf{Landmark Error:} This metric assesses the accuracy of the mapping by comparing it against ground truth landmark correspondences. Given a set of landmark pairs $(p_i, q_i)$, where $p_i \in \mathcal{S}$ and $q_i \in \mathcal{T}$, the matching error is defined as:
    \begin{equation}
        \text{Landmark Error}(f) = \frac{1}{n} \sum_{i=1}^n \frac{d(f(p_i), q_i)}{\sqrt{area(\mathcal{T})}},
    \end{equation}
    where $d(\cdot, \cdot)$ is the geodesic distance, $n$ is the number of landmark pairs and $\sqrt{area(\cdot)}$ is a normalizing factor. 
    A lower matching error indicates a more accurate mapping.
    
\noindent\textbf{Bijectivity:}
Let \( \mathcal{S} \) denote the source point cloud and \( \mathcal{T} \) denote the target point cloud. We define \( M_{12} \) as the mapping from \( \mathcal{S} \) to \( \mathcal{T} \) and \( M_{21} \) as the mapping from \( \mathcal{T} \) to \( \mathcal{S} \). Furthermore, let \( M_{11} = M_{12}(M_{21}) \) and \( M_{22} = M_{21}(M_{12}) \). The transformation error can be quantified using the following expressions:

\begin{align}
\text{V}_S=\text{vec}(\text{normv}(\mathcal{S} - \mathcal{S}[M_{11}])) \\
\text{V}_T=\text{vec}(\text{normv}(\mathcal{T} - \mathcal{T}[M_{22}])) \\
\text{Bijectivity}=\frac{\frac{1}{n} \sum_{i=1}^{n} \mathbf{V}_S^i + \frac{1}{m} \sum_{j=1}^{m} \mathbf{V}_T^j}{2},
\end{align}
where \( \mathcal{S}[M_{11}] \) represents the points in \( \mathcal{S} \) after applying the mapping \( M_{11} \), and \( \mathcal{T}[M_{22}] \) represents the points in \( \mathcal{T} \) after applying the mapping \( M_{22} \). Here, \(\text{vec}(\cdot)\) denotes the vectorization operation, and \(\text{normv}(\cdot)\) denotes the computation of the norm of the vectors.

\begin{figure}[h!]
  \begin{center}
\includegraphics[width=8cm]{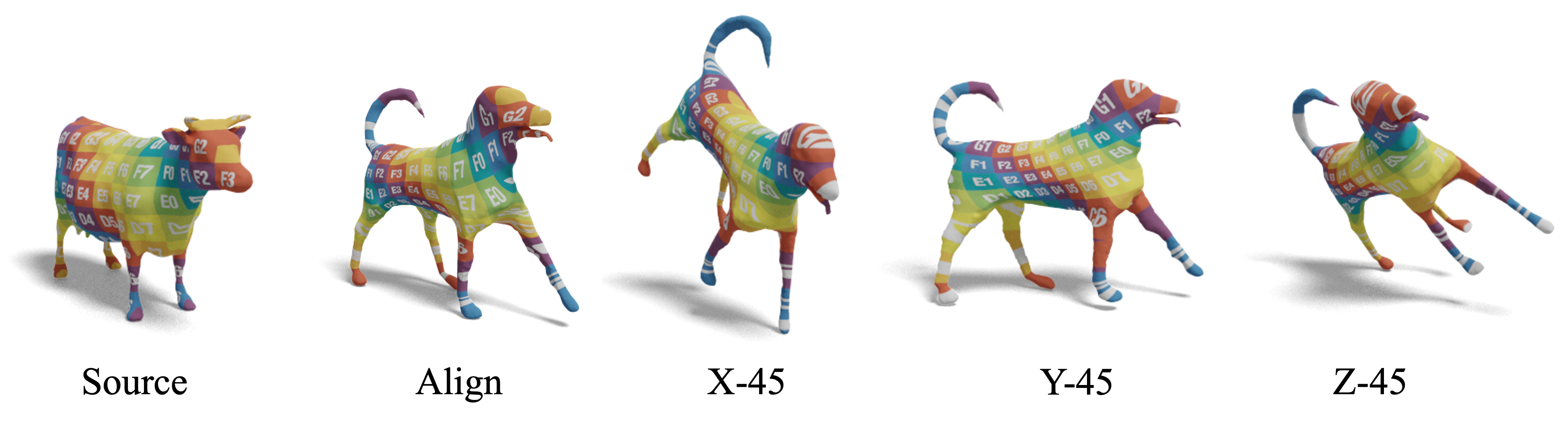}
  \end{center}
\caption{We initially rotate the target shape by 45 degrees along the X, Y, and Z axes, respectively. The texture visualization indicates that our method is relatively robust to rotation.}
\label{fig:rotation}
\end{figure}

\subsection{Robustness regarding misalignment}\label{supp:rot}

We provide visualizations corresponding to Tab.3 of the main paper as shown in Fig.~\ref{fig:rotation}. 
Fixing the source shape, we rotate the target shape around the X, Y, and Z axes by 45 degrees, respectively. 
Thanks to the strong robustness of LVM regarding rotational perturbations in images, we can obtain reasonable image interpolation results, which can provide appropriate guidance during the registration process. 
Fig.~\ref{fig:rotation} demonstrates that, without explicitly optimizing for rotation, our method maintains good robustness to rotations of up to 45 degrees.

\begin{figure}[b!]
  \begin{center}
\includegraphics[width=8cm]{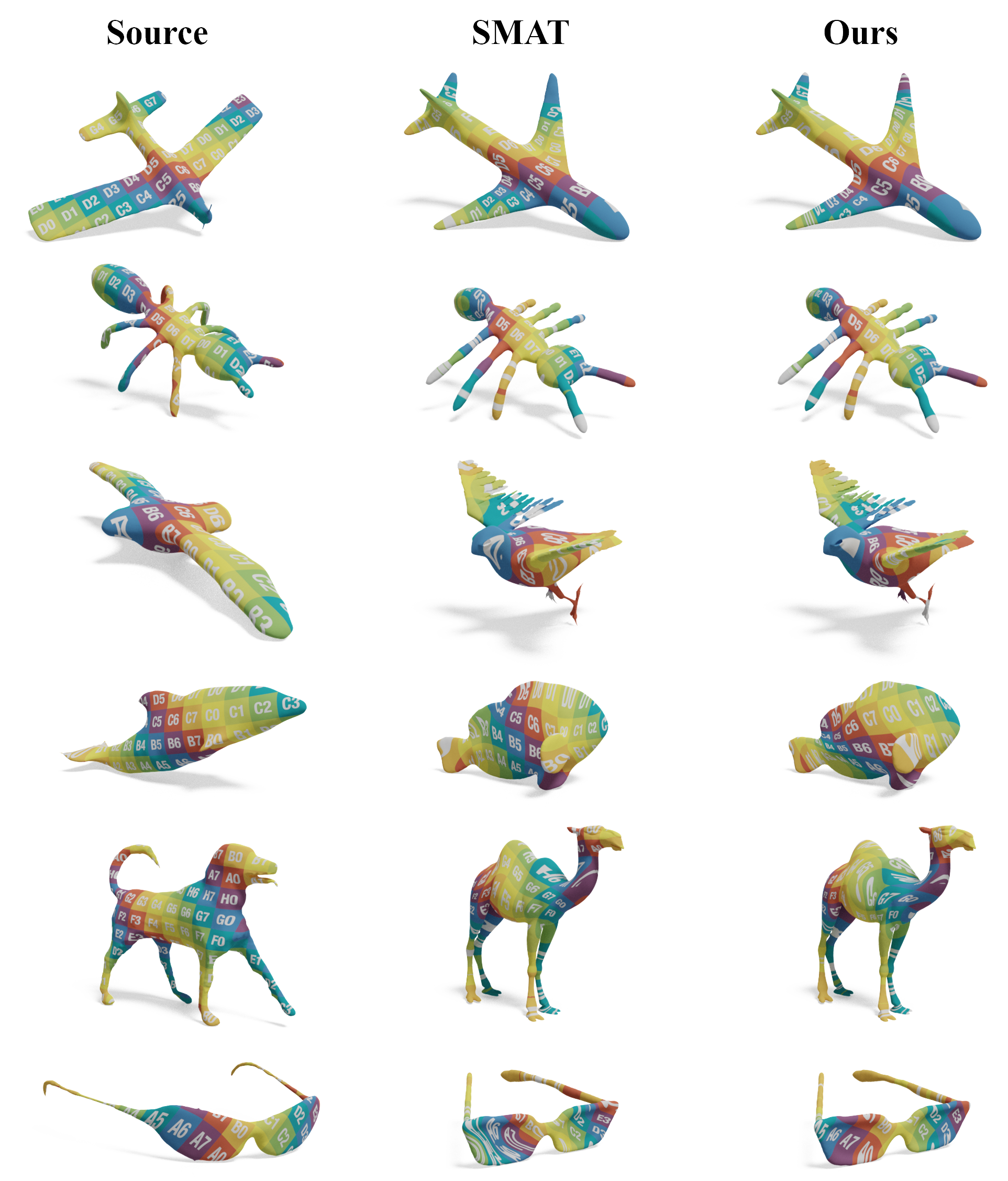}
  \end{center}
\caption{Qualitative comparisons between \textbf{SRIF} and SMAT. Note the latter consumes ground-truth landmarks as input, while \textbf{SRIF} is fully automatic. }
\label{fig:smat}
\end{figure}

\subsection{Qualitative comparison to SMAT}\label{supp:smat}

To further evaluate the plausibility of our results, we apply SMAT on 6 challenging pairs from Fig. 10 of the main paper in Fig.~\ref{fig:smat} and compare the results with ours. 
Note that we feed in all available landmarks to SMAT, and that our method achieves comparable results with SMAT while being fully automatic.

\subsection{Non-rigid shape matching}\label{supp:human}

\begin{figure}[t!]
  \begin{center}
\includegraphics[width=9cm]{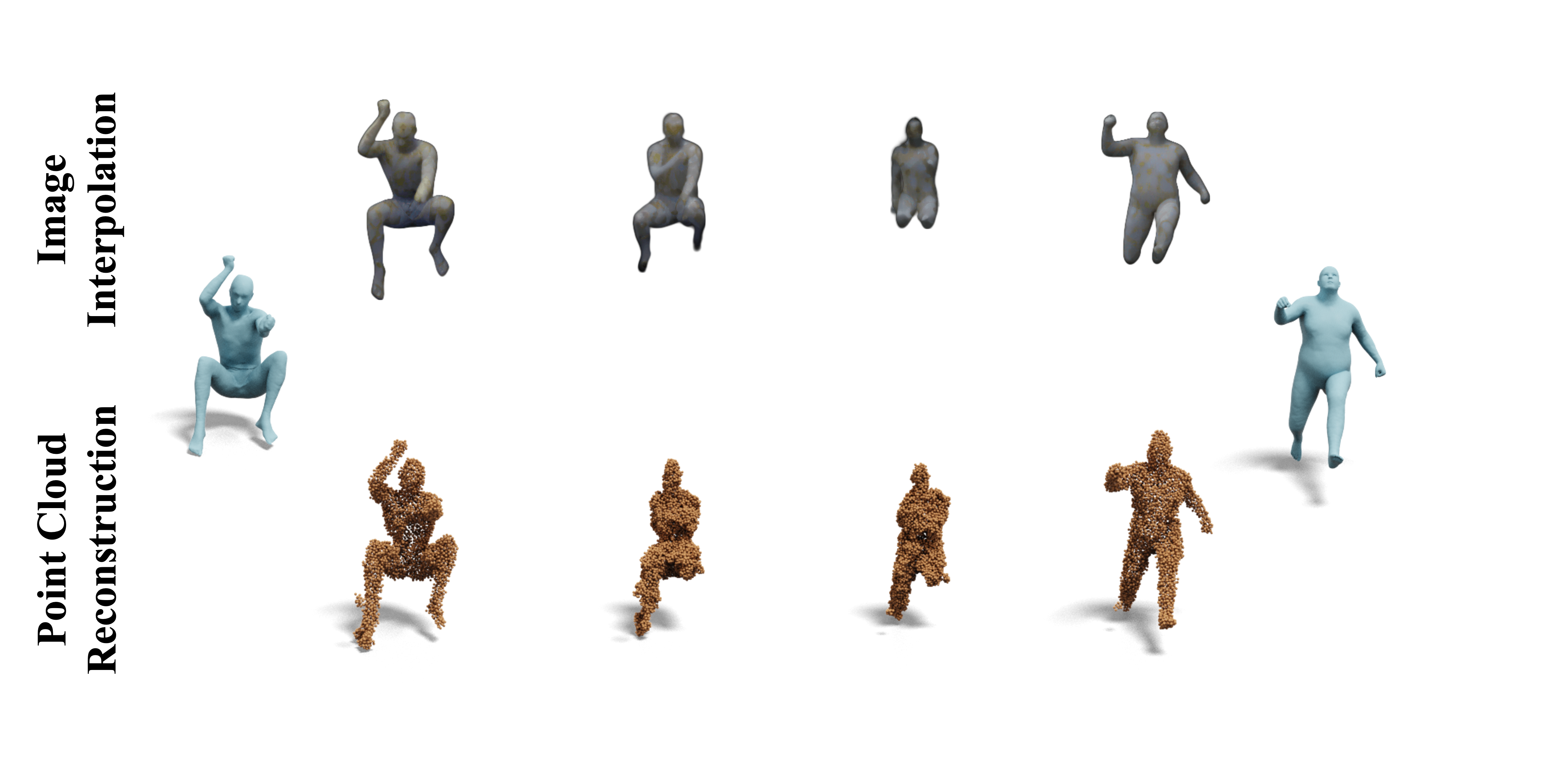}
  \end{center}
\caption{We evaluate our method on a pair of human shapes undergoing large non-rigid deformations. Top row: Image interpolation; Bottom row: Intermediate point cloud reconstructions.}
\label{fig:failure}
\end{figure}

In this section, we discuss the performance of our method on datasets with significant pose variations. 
We selected the following two datasets:  
For SHREC19~\cite{SHREC19}, we evaluate 407 pairs of data with ground truth. In particular, we exclude 23 pairs related to a partial shape. 
For SMPL~\cite{SMPL}, we randomly generate a set of 500 shapes using SMPL model. And then sample 20 shapes from them via FPS in the pose parameters of generation.
Subsequently, similar to the procedure in Sec.4.2 of the main text, we construct 50 pairs among the 20 shapes for inference. As shown in Tab. 5 of the main text, our method is outperformed by SmoothShells with a noticeable margin. 

To investigate the failure cause, we examine the intermediate results of our pipeline. 
As shown in Fig.~\ref{fig:failure}, when there are significant differences in the pose, the image interpolation module~\cite{diffmorpher} struggles to return plausible results (top row), which is further amplified in the follow-up point cloud reconstruction (bottom row). 
Such discrepancy then leads to the suboptimal solution of our method on this benchmark.

\subsection{Running Time analysis}\label{supp:time}

\begin{figure}[t!]
  \begin{center}
\includegraphics[width=8cm]{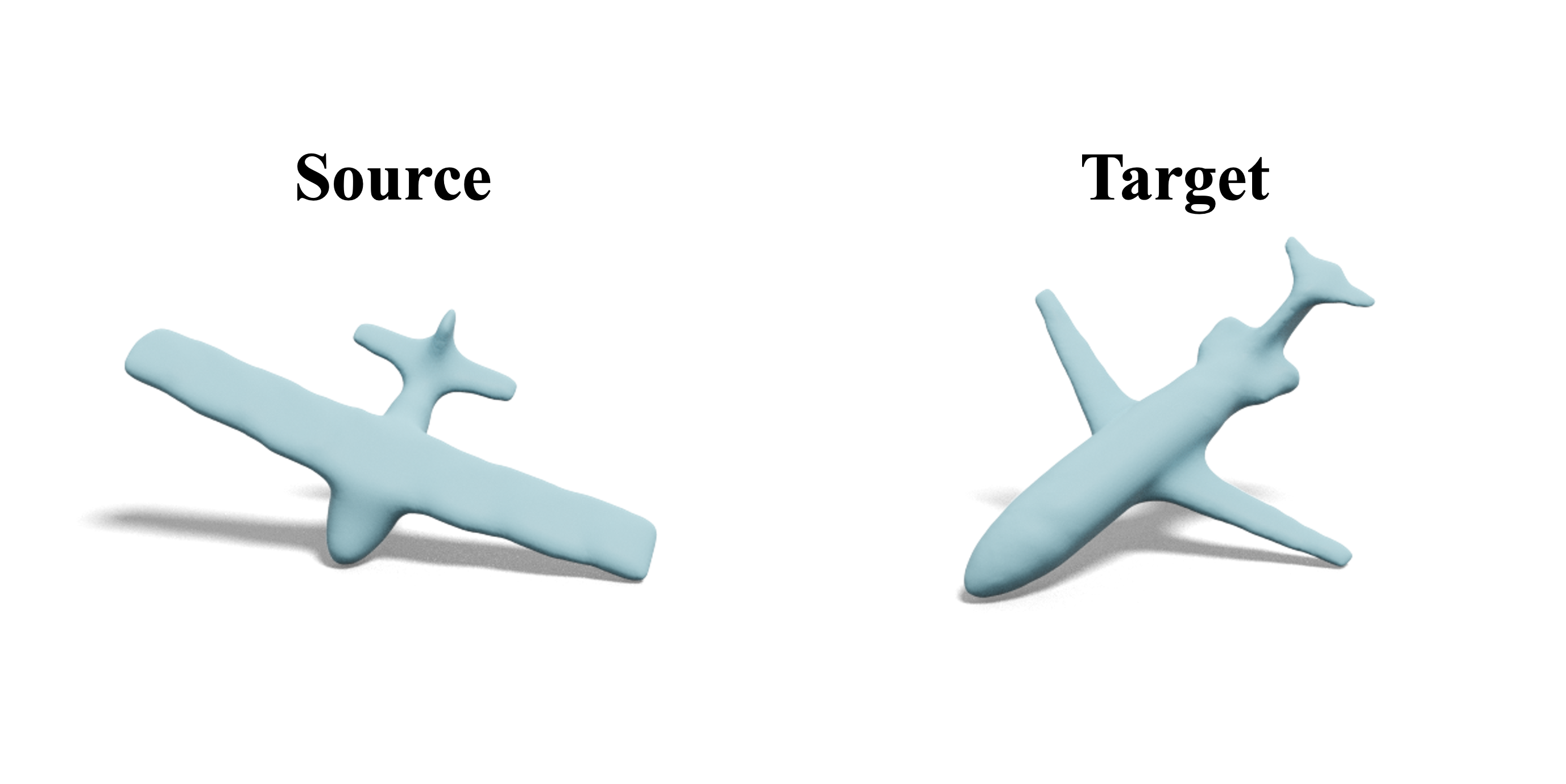}
  \end{center}
\caption{We test the running time of all the methods on this pair.}
\label{fig:runtime}
\end{figure}

We test MapTree, SmoothShells, NeuroMorph, BIM and our method on the same device, which includes an NVIDIA V100 GPU, a single-core 2.8GHz CPU, and 500MB of memory. 
The test pair is shown in Fig~\ref{fig:runtime}, where the source mesh contains $5400$ points and the target mesh contains $5619$ points. 

In particular, we evaluate MapTree~\footnote{https://github.com/llorz/SGA20\_mapExplor}, SmoothShells~\footnote{https://github.com/marvin-eisenberger/smooth-shells}, NeuroMorph~\footnote{https://github.com/facebookresearch/neuromorph}, and BIM on all the test data with the regarding official implementations. 
Regarding ENIGMA, since the code is not publicly available, we asked the authors to run baseline evaluation, who also reported that ENIGMA took 20 mins to match shapes of 5000 vertices with RHM~\cite{rhm} as post-refinement.

\subsection{Per-category result analysis}\label{supp:percate}
We report the scores regarding the four metrics in Sec.~\ref{supp:metric} for each category in Tab.~\ref{table:full_results} as a supplement to Tab.1 in the main paper. 

\begin{table*}[ht]
\caption{Quantitative results of the test cases shown in Fig.8 of the main submission. SS stands for SmoothShell, and NM stands for NeuroMorph. For Enigma, we follow their symmetric setup by calculating the error of both forward and reverse mapping in the SHREC07 categories and taking the minimum of these two as the final result. The former and latter landmarks errors correspond to the outcomes of the forward mapping and the symmetric result, respectively. 
}\label{table:full_results}
\centering
\begin{tabular}{
>{\columncolor[HTML]{FFFFFF}}c 
>{\columncolor[HTML]{FFFFFF}}c 
>{\columncolor[HTML]{FFFFFF}}c 
>{\columncolor[HTML]{FFFFFF}}c 
>{\columncolor[HTML]{FFFFFF}}c 
>{\columncolor[HTML]{FFFFFF}}c 
>{\columncolor[HTML]{FFFFFF}}c 
>{\columncolor[HTML]{FFFFFF}}c 
>{\columncolor[HTML]{FFFFFF}}c 
>{\columncolor[HTML]{FFFFFF}}c 
>{\columncolor[HTML]{FFFFFF}}c 
>{\columncolor[HTML]{FFFFFF}}c }
\hline
                                                   &             & Dirichlet ↓ & Cov. ↑        & Lmks. Err. ↓                 & Bij ↓             &                                                   &             & Dirichlet ↓ & Cov. ↑         & Lmks. Err. ↓         & Bij ↓              \\ \hline
\cellcolor[HTML]{FFFFFF}                           & ENIGMA      & 2.5703            & 0.6716          & 0.35\textbackslash{}0.25 & 0.0059          & \cellcolor[HTML]{FFFFFF}                          & ENIGMA      & 8.3737  & 0.6429 & 0.37\textbackslash{}0.25 & 0.0069 \\
\cellcolor[HTML]{FFFFFF}                           & Maptree     & 14.1676           & 0.3633          & 0.5634                   & 0.0487          & \cellcolor[HTML]{FFFFFF}                          & Maptree     & 13.2092           & 0.2680           & 0.2635           & 0.0589           \\
\cellcolor[HTML]{FFFFFF}                           & BIM         & 3.4598            & 0.6525          & 0.2589                   & 0.2223          & \cellcolor[HTML]{FFFFFF}                          & BIM         & 8.5061            & 0.6230           & 0.3050           & 0.2347           \\
\cellcolor[HTML]{FFFFFF}                           & SS & 7.3184            & 0.7362          & 0.2749                   & 0.0064          & \cellcolor[HTML]{FFFFFF}                          & SS & 12.0404           & \textbf{0.7430}  & 0.2694           & 0.0073           \\
\cellcolor[HTML]{FFFFFF}                           & NM  & 8.1679            & 0.1127          & 0.1510                   & 0.0831          & \cellcolor[HTML]{FFFFFF}                          & NM  & 16.0912           & 0.0842           & 0.1895           & 0.0940           \\
\multirow{-6}{*}{\cellcolor[HTML]{FFFFFF}airplane} & Ours        & \textbf{2.3427}   & \textbf{0.7557} & \textbf{0.0356}          & \textbf{0.0039} & \multirow{-6}{*}{\cellcolor[HTML]{FFFFFF}ant}     & Ours        & \textbf{7.5907}   & 0.6868           & \textbf{0.1133}  & \textbf{0.0068}  \\ \hline
\cellcolor[HTML]{FFFFFF}                           & ENIGMA      & \textbf{4.1129}   & 0.5529          & 0.45\textbackslash{}0.31 & 0.0075          & \cellcolor[HTML]{FFFFFF}                          & ENIGMA      & 8.2018  & 0.3350 & 0.33\textbackslash{}0.28 & 0.0153 \\
\cellcolor[HTML]{FFFFFF}                           & Maptree     & 19.9621           & 0.3208          & 0.4742                   & 0.0554          & \cellcolor[HTML]{FFFFFF}                          & Maptree     & 9.7311            & 0.2110           & 0.4117           & 0.0564           \\
\cellcolor[HTML]{FFFFFF}                           & BIM         & 32.2091           & 0.3579          & 0.4800                   & 0.3160          & \cellcolor[HTML]{FFFFFF}                          & BIM         & 8.3057            & 0.3333           & 0.4123           & 0.2888           \\
\cellcolor[HTML]{FFFFFF}                           & SS & 28.3655           & 0.5755          & 0.2627                   & 0.0321          & \cellcolor[HTML]{FFFFFF}                          & SS & 26.5909           & 0.3694           & 0.3009           & 0.0193           \\
\cellcolor[HTML]{FFFFFF}                           & NM  & 18.0280           & 0.0266          & 0.1853                   & 0.0922          & \cellcolor[HTML]{FFFFFF}                          & NM  & 13.5221           & 0.0895           & 0.1672           & 0.1178           \\
\multirow{-6}{*}{\cellcolor[HTML]{FFFFFF}chair}    & Ours        & 5.6372            & \textbf{0.6127} & \textbf{0.0295}          & \textbf{0.0052} & \multirow{-6}{*}{\cellcolor[HTML]{FFFFFF}bird}    & Ours        & \textbf{5.8626}   & \textbf{0.5880}  & \textbf{0.0965}  & \textbf{0.0121}  \\ \hline
\cellcolor[HTML]{FFFFFF}                           & ENIGMA      & \textbf{2.3935}   & 0.7360          & 0.17\textbackslash{}0.14 & 0.0055          & \cellcolor[HTML]{FFFFFF}                          & ENIGMA      & 5.6478  & 0.7372 & 0.68\textbackslash{}0.32 & 0.0175 \\
\cellcolor[HTML]{FFFFFF}                           & Maptree     & 10.2440           & 0.6344          & 0.2949                   & 0.0271          & \cellcolor[HTML]{FFFFFF}                          & Maptree     & 9.0106            & 0.5915           & 0.6878           & 0.0376           \\
\cellcolor[HTML]{FFFFFF}                           & BIM         & 4.2617            & 0.6656          & 0.2699                   & 0.1567          & \cellcolor[HTML]{FFFFFF}                          & BIM         & 31.3520           & 0.5013           & 0.6134           & 0.4158           \\
\cellcolor[HTML]{FFFFFF}                           & SS & 4.7195            & \textbf{0.7927} & 0.1032                   & 0.0048          & \cellcolor[HTML]{FFFFFF}                          & SS & 9.7207            & 0.7547           & 0.3100           & 0.0177           \\
\cellcolor[HTML]{FFFFFF}                           & NM  & 13.9717           & 0.1584          & 0.1366                   & 0.0787          & \cellcolor[HTML]{FFFFFF}                          & NM  & 9.2590            & 0.0753           & 0.2652           & 0.1437           \\
\multirow{-6}{*}{\cellcolor[HTML]{FFFFFF}fish}     & Ours        & 3.1247            & 0.7668          & \textbf{0.0804}          & \textbf{0.0040} & \multirow{-6}{*}{\cellcolor[HTML]{FFFFFF}glasses} & Ours        & \textbf{5.6394}   & \textbf{0.7580}  & \textbf{0.0614}  & \textbf{0.0102}  \\ \hline
\cellcolor[HTML]{FFFFFF}                           & ENIGMA      & \textbf{4.5125}   & 0.3133          & 0.25\textbackslash{}0.19 & 0.0130          & \cellcolor[HTML]{FFFFFF}                          & ENIGMA      & 16.5595  & 0.7656 & 0.44\textbackslash{}0.31 & 0.0145 \\
\cellcolor[HTML]{FFFFFF}                           & Maptree     & 11.0492           & 0.2260          & 0.3507                   & 0.0426          & \cellcolor[HTML]{FFFFFF}                          & Maptree     & 34.3151           & 0.4657           & 0.2805           & 0.0371           \\
\cellcolor[HTML]{FFFFFF}                           & BIM         & 8.8194            & 0.2846          & 0.2449                   & 0.2010          & \cellcolor[HTML]{FFFFFF}                          & BIM         & 15.3346           & 0.7045           & 0.5197           & 0.3149           \\
\cellcolor[HTML]{FFFFFF}                           & SS & 8.8495            & 0.3233          & 0.1234                   & 0.0116          & \cellcolor[HTML]{FFFFFF}                          & SS & 19.0768           & \textbf{0.8010}  & 0.3900           & 0.0081           \\
\cellcolor[HTML]{FFFFFF}                           & NM  & 18.4938           & 0.0692          & 0.1697                   & 0.1070          & \cellcolor[HTML]{FFFFFF}                          & NM  & 36.0034           & 0.1251           & 0.2830           & 0.1055           \\
\multirow{-6}{*}{\cellcolor[HTML]{FFFFFF}fourleg}  & Ours        & 6.2250            & \textbf{0.3512} & \textbf{0.0824}          & \textbf{0.0114} & \multirow{-6}{*}{\cellcolor[HTML]{FFFFFF}plier}   & Ours        & \textbf{13.2623}  & 0.7508           & \textbf{0.1476}  & \textbf{0.0080}  \\ \hline
\cellcolor[HTML]{FFFFFF}                           & ENIGMA      & \textbf{4.8419}   & \textbf{0.8108} & 0.29\textbackslash{}0.14 & \textbf{0.0075} & \cellcolor[HTML]{FFFFFF}                          & ENIGMA      & \textbf{5.8304}   & 0.6026           & 0.3032           & 0.0060           \\
\cellcolor[HTML]{FFFFFF}                           & Maptree     & 14.6992           & 0.3532          & 0.2633                   & 0.0373          & \cellcolor[HTML]{FFFFFF}                          & Maptree     & 26.667            & 0.5331           & 0.4231           & 0.031            \\
\cellcolor[HTML]{FFFFFF}                           & BIM         & 7.1003            & 0.5037          & 0.1810                   & 0.1205          & \cellcolor[HTML]{FFFFFF}                          & BIM         & 5.3741            & 0.5456           & 0.2968           & 0.2305           \\
\cellcolor[HTML]{FFFFFF}                           & SS & 8.7462            & 0.5304          & 0.1572                   & 0.0164          & \cellcolor[HTML]{FFFFFF}                          & SS & 14.7793           & 0.6495           & 0.4476           & 0.0059           \\
\cellcolor[HTML]{FFFFFF}                           & NM  & 28.5119           & 0.0798          & 0.2141                   & 0.0997          & \cellcolor[HTML]{FFFFFF}                          & NM  & 59.3832           & 0.2509           & 0.1844           & 0.0437           \\
\multirow{-6}{*}{\cellcolor[HTML]{FFFFFF}human}    & Ours        & 8.6968            & 0.5394          & \textbf{0.1467}          & 0.0100          & \multirow{-6}{*}{\cellcolor[HTML]{FFFFFF}ebcm}    & Ours        & 6.3403            & \textbf{0.6784}  & \textbf{0.0886}  & \textbf{0.0038}  \\ \hline
\end{tabular}
\end{table*}
Considering symmetry in the results for ENIGMA, our outcomes are superior in four out of five categories compared to ENIGMA.

\begin{figure*}[htpb!]
  \begin{center}
\includegraphics[width=17.5cm]{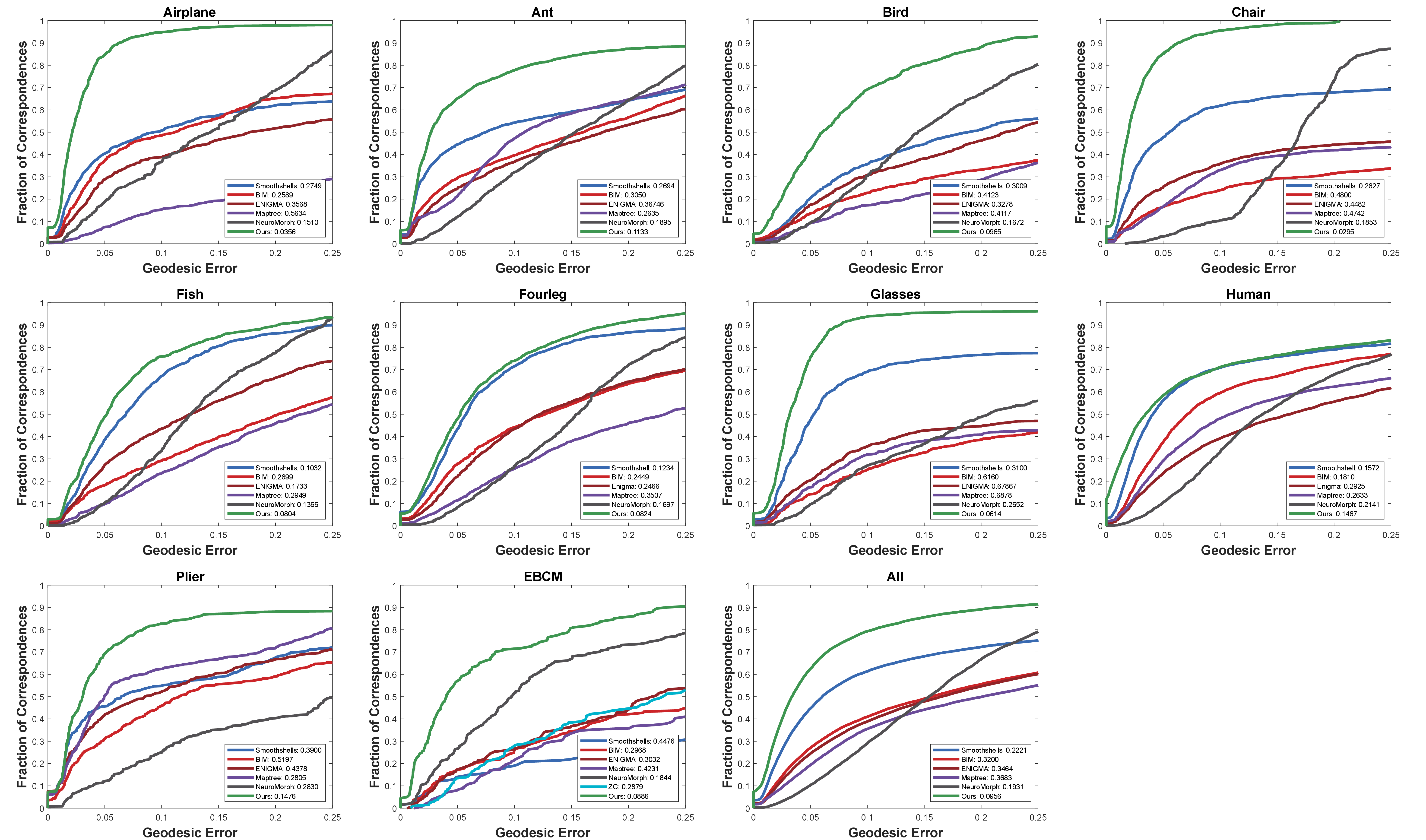}
  \end{center}
\caption{Accuracy evaluation of our method and baselines on 10 test sets. The curves read the fraction (Y-axis) of computed correspondences that fall within certain normalized geodesic distance to the ground-truth ones (X-axis). The numbers in the legends show the average error. Our method achieves the best accuracy over \emph{all} sets.}
\label{fig:results}
\end{figure*}

We also provide accumulated error curves of the ten categories in Fig.~\ref{fig:results}. 
It is evident that in the categories of airplane, chair, ant, bird, glasses, plier, and EBCM, our method's landmark error reduction is prominent. 
Indeed, our method gains at least \textbf{$40\%$} improvement upon the second-best results. 

\begin{figure*}[h!]
  \begin{center}
\includegraphics[width=17.5cm]{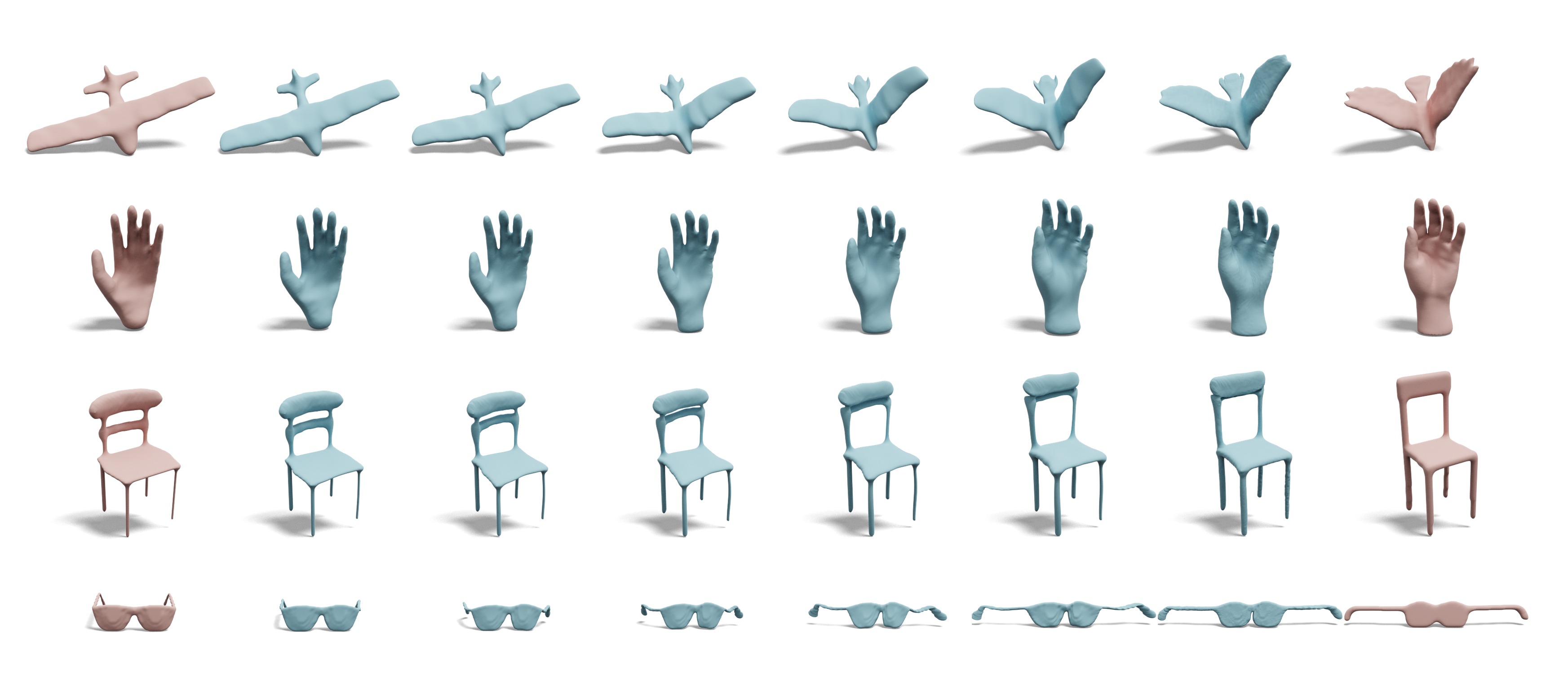}
  \end{center}
\caption{Demonstrations of smooth and semantically meaningful shape interpolations obtained by our estimated flow. }
\label{fig:interpolation_supp}
\end{figure*}

We also provide more shape morphing results in Fig.~\ref{fig:interpolation_supp}.

\bibliographystyle{ACM-Reference-Format}
\bibliography{sample-bibliography}

\end{document}